%% file: imf.tex
\documentclass[10pt,twocolumn,letterpaper]{article}

\PassOptionsToPackage{numbers}{natbib}

\usepackage[pagenumbers]{cvpr}
\setcitestyle{numbers,square}
\definecolor{cvprblue}{rgb}{0.21,0.49,0.74}
\usepackage[breaklinks,colorlinks,allcolors=cvprblue]{hyperref}
\usepackage{tikz} 
\usepackage[utf8]{inputenc} 
\usepackage[T1]{fontenc} 
\usepackage{url} 
\usepackage{microtype} 
\usepackage{algorithm}
\usepackage{indentfirst}
\usepackage{algpseudocode}
\usepackage{listings}
\usepackage{overpic}
\usepackage{wrapfig}
\usepackage{colortbl}
\usepackage{tabularx}
\usepackage{verbatim}
\usepackage{nicefrac}
\usepackage{empheq} 
\usepackage{times}
\usepackage{booktabs}
\usepackage{caption}
\usepackage{subcaption}
\usepackage{multirow}
\usepackage{makecell}
\usepackage{adjustbox}
\usepackage{graphicx}
\usepackage{array}

\usepackage{soul}
\usepackage{tikz}
\usepackage{amsmath} 
\usepackage{pifont}

\usepackage{centernot}
\usepackage{xcolor}
\usepackage{amsthm}
\usepackage{thmtools}
\usepackage{thm-restate}
\definecolor{citecolor}{HTML}{0071BC}
\definecolor{linkcolor}{HTML}{ED1C24}

\usetikzlibrary{arrows.meta, positioning, calc}

\definecolor{myorange}{RGB}{255,127,0} 
\definecolor{myblue}{RGB}{0,114,189} 
\def\paperID{8640} 
\def\confName{CVPR}
\def\confYear{2026}

\usepackage{amssymb}
\usepackage{amsmath}
\usepackage{amsfonts} 
\usepackage{nicefrac} 
\usepackage{mathtools}

\input{macro}

\title{
    Improved Mean Flows: 
    On the Challenges of Fastforward Generative Models
}

\newcommand{\authorskip}{\hspace{1.8mm}}
\author{%
Zhengyang Geng$^{1,2,3,}$\thanks{Equal contribution. Part of this work was done when Z. Geng was interning at Adobe and MIT, and when Y. Lu was interning at MIT.}
\authorskip Yiyang Lu$^{4,2,*}$
\authorskip Zongze Wu$^{3}$
\authorskip Eli Shechtman$^{3}$
\authorskip J. Zico Kolter$^{1}$
\authorskip Kaiming He$^{2}$\\
\\
$^{1}$CMU \quad $^{2}$MIT \quad $^{3}$Adobe \quad $^{4}$THU }

\begin{document}
\maketitle

\input{sections/abstract}
\input{sections/introduction}

\input{sections/related_works}
\input{sections/background}

\input{sections/method}
\input{sections/experiments}

\input{sections/conclusion}
\input{sections/appendix}
\input{sections/acknowledgement}

\renewcommand{\topfraction}{0.9}
\renewcommand{\bottomfraction}{0.9}
\renewcommand{\textfraction}{0.1}
\renewcommand{\floatpagefraction}{0.9}
\setcounter{topnumber}{5}
\setcounter{bottomnumber}{5}
\setcounter{totalnumber}{10}

\input{sections/vis}

{
    \small
    \bibliographystyle{ieeenat_fullname}
    \bibliography{imf}
}
\end{document}

%% file: macro.tex
\usepackage[capitalize]{cleveref}
\usepackage{subcaption}

\crefname{equation}{Eq.}{Eq.}
\crefname{figure}{Fig.}{Fig.}
\crefname{table}{Tab.}{Tab.~}
\crefname{section}{Sec.}{Sec.~}
\crefname{algorithm}{Alg.}{Alg.~}
\crefname{thm}{Theorem}{Theorem~}
\crefname{lemma}{Lemma}{Lemma~}
\crefname{appendix}{Appendix}{Appendix~}

\theoremstyle{plain}

\theoremstyle{definition}

\theoremstyle{remark}

\def\ie{\textit{i.e.,~}}
\def\eg{\textit{e.g.,~}}

\def\wrt{\textit{w.r.t.~}}

\newcommand{\ddt}{\frac{d}{d t}}

\newcommand{\tablestyle}[2]{\setlength{\tabcolsep}{#1}\renewcommand{\arraystretch}{#2}\centering\footnotesize}

\newcolumntype{x}[1]{>{\centering\arraybackslash}p{#1pt}}
\newcolumntype{y}[1]{>{\raggedright\arraybackslash}p{#1pt}}
\newcolumntype{z}[1]{>{\raggedleft\arraybackslash}p{#1pt}}

\definecolor{selectcolor}{gray}{.85}

\renewcommand{\paragraph}[1]{\vspace{1.25mm}\noindent\textbf{#1}}

\newcommand{\utgt}{u_\text{tgt}}

\newcommand\highlight[1]{{\color{red}#1}}
\newcommand\deemph[1]{{\textcolor[gray]{0.6}{#1}}}

\newcommand{\sgjvp}{\mathtt{JVP}_\textrm{sg}}
\newcommand{\app}{\raise.17ex\hbox{$\scriptstyle\sim$}}

%% file: sections/abstract.tex
\begin{abstract}
MeanFlow (MF) has recently been established as a framework for one-step generative modeling. However, its ``fastforward'' nature introduces key challenges in both the training objective and the guidance mechanism.
First, the original MF's training target depends not only on the underlying ground-truth fields but also on the network itself. To address this issue, we recast the objective as a loss on the instantaneous velocity $v$, re-parameterized by a network that predicts the average velocity $u$.
Our reformulation yields a more standard regression problem and improves the training stability.
Second, the original MF fixes the classifier-free guidance scale during training, which sacrifices flexibility.
We tackle this issue by formulating guidance as explicit conditioning variables, thereby retaining flexibility at test time.
The diverse conditions are processed through in-context conditioning, which reduces model size and benefits performance.
Overall, our \textbf{improved MeanFlow} (\textbf{iMF}) method, trained entirely from scratch, achieves \textbf{1.72} FID with a single function evaluation (1-NFE) on ImageNet 256$\times$256. 
iMF substantially outperforms prior methods of this kind and closes the gap with multi-step methods while using no distillation. 
We hope our work will further advance fastforward generative modeling as a stand-alone paradigm.
\end{abstract}

%% file: sections/introduction.tex
\vspace{-1.em}
\section{Introduction}

Diffusion models~\cite{diffusion,ddpm,scoresde} and their flow-based variants~\cite{fm,rectified,stochasticflow} are highly effective for generative modeling.
These models can be viewed as solving a differential equation (\eg an ODE) that maps a prior distribution to the data distribution.
Because these equations are typically solved using multi-step numerical solvers, the generation process requires a certain number of function evaluations (NFEs).

Recently, encouraging progress~\cite{cm,ict,ect,scm,shortcut,imm,mf} has been made toward reducing sampling steps in diffusion/flow-based models.
Using the concept from physical simulation, these models can be thought of as \emph{fastforward} approximations to the underlying differential equations.
The resulting fastforward models are capable of generating in a very few or even one step.
To achieve this goal, the training objectives are formulated as look-ahead mappings that operate across large time intervals, and various approximations have been proposed to address this challenging problem.

\input{sections/teaser}

In this work, we take a deeper look at the recently proposed MeanFlow (MF) framework~\cite{mf}.
In MF, instead of learning the \emph{instantaneous velocity} field (denoted by $v$) underlying the ODE, the model learns an \emph{average velocity} field (denoted by $u$) across time steps. 
To avoid infeasible integration during training, MF reformulates the problem into a differential relation between the instantaneous and average velocity fields.
This relation, called the ``MeanFlow identity''~\cite{mf}, establishes a trainable objective.  
The underlying average velocity field serves as the ground-truth and as the optimum of this training objective.

Despite the encouraging results of the original MF~\cite{mf}, we identify two major issues that remain unresolved: (i) the training target in the original MF is network-dependent and therefore does not constitute a standard regression problem; (ii) MF handles the classifier-free guidance (CFG) \cite{cfg} using a fixed training-time guidance scale, which sacrifices flexibility. 
We analyze these issues and present our solutions.

First, the original MF predicts the average velocity $u$, an unknown quantity that is substituted with the network's own prediction $u_\theta$. To have a network-agnostic prediction target, we show that the original MF can be equivalently reformulated as a loss on the instantaneous velocity (namely, $v$-loss), which is \textit{re-parameterized} by the network that predicts the average velocity $u$ (namely, $u$-pred). See \cref{fig:teaser}\textbf{(a)}. 
This reformulation provides a regression target $v$ that does not depend on the network.
From this perspective, we further propose to reformulate the regression input, enforcing it to depend only on the noisy sample but not on other unknown quantities (\cref{fig:teaser}\textbf{(b)}). Our improved objective substantially stabilizes the training process in practice.

Second, the original MF handles CFG \cite{cfg} using a \textit{fixed} guidance scale that is determined before training. We suggest that fixing the guidance compromises the flexibility at inference time, and that the optimal value depends on the model’s capability. To address this, we reformulate the guidance as a form of \textit{conditioning}, allowing it to take varying values during both training and inference. This formulation unlocks the power and flexibility of CFG while still maintaining the 1-NFE sampling behavior.
We further design an improved architecture that accommodates this and other types of conditions through in-context conditioning.

Overall, our experiments show that our \textbf{improved MeanFlow} (\textbf{iMF}) effectively addresses these issues in the original MF. In the challenging setting of 1-NFE generation on ImageNet 256$\times$256 trained from scratch, iMF achieves an FID of \textbf{1.72}, representing a relative 50\% improvement over the original MF and setting a new state-of-the-art of its kind.
Our models do not use distillation or any pre-trained models for alignment. 
This result substantially narrows the gap with those of multi-step methods, suggesting that fastforward generative models can be a promising stand-alone framework.

%% file: sections/teaser.tex
\begin{figure}
    \centering
    \vspace{-1em}
    \begin{subfigure}[t]{0.35\linewidth}
        \centering
        \includegraphics[width=\linewidth,clip,page=1]{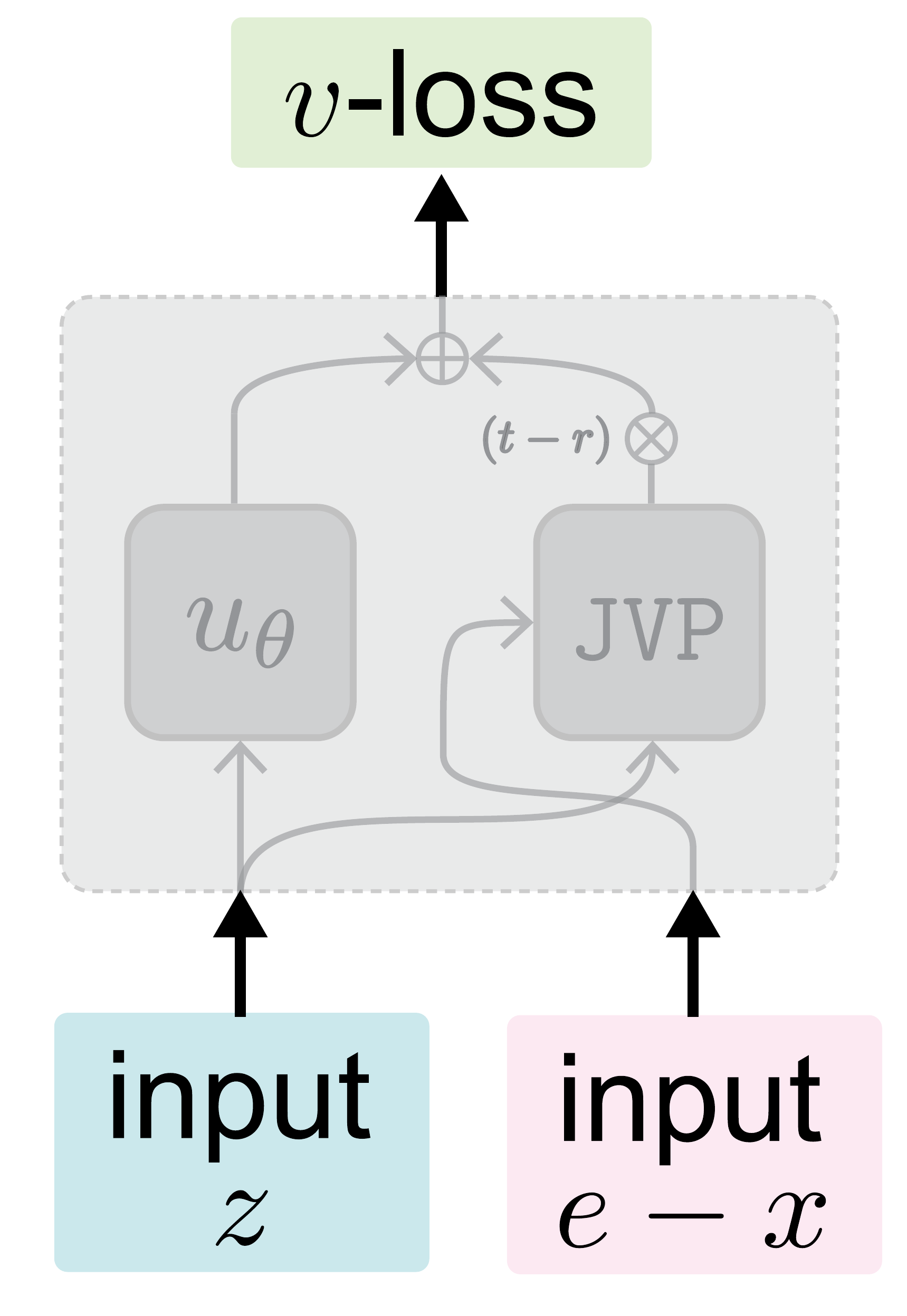}
        \caption{\centering Original MeanFlow}
        \label{fig:teaser_mf}
    \end{subfigure}
    \hspace{0.1\linewidth}
    \begin{subfigure}[t]{0.35\linewidth}
        \centering
        \includegraphics[width=\linewidth,clip,page=2]{figs/teaser.pdf}
        \caption{\centering Improved MeanFlow}
        \label{fig:teaser_imf}
    \end{subfigure}
    \vspace{-.2em}
    \caption{
    \textbf{Conceptual comparison}. Original MeanFlow (MF) \cite{mf} predicts \textit{average velocity} $u$ by a network $u_\theta$. As the ground-truth $u$ is unknown, original MF substitutes $u$ with the network's own prediction. We show that the original MF objective is equivalent to a loss on the \textit{instantaneous velocity} $v$ (namely, $v$-loss), but re-parameterized by the neural network $u_\theta$ (namely, $u$-pred), as shown in \textbf{(a)}. This re-parameterization, encompassed within the gray box, is determined by \textit{the MeanFlow identity} \cite{mf}.
    This reformulation reveals that the input to the compound function (in the gray box) is not only the noisy data (here, $z$), but also the conditional velocity ($e - x$), which is not a standard regression problem. In \textbf{(b)}, our improved objective is conceptually \emph{\textbf{$v$-loss re-parameterized by $u$-pred}}, taking only the legitimate input $z$.
    }
    \vspace{-.7em}
    \label{fig:teaser}
\end{figure}

%% file: sections/related_works.tex
\section{Related Work}

\paragraph{Diffusion and Flow-based Models.}
Diffusion models~\cite{diffusion,ddpm,edm,ncsn,scoresde} and flow matching~\cite{fm,rectified,stochastic} lay the foundation for a series of modern generative methods. 
These approaches can be formulated as learning a probabilistic trajectory, \ie an ODE/SDE (ordinary/stochastic differential equation) that maps between distributions.
A network is trained to model the underlying trajectory using a regression loss.
Samples are generated by solving the resulting ODE or SDE, typically using a numerical solver.

\paragraph{Fastforward Generative Models.}
Standard diffusion and flow-based models were originally designed without explicitly considering the acceleration of ODE/SDE solving.
An emerging category of methods, which we abstract as ``fastforward generative models'', explicitly incorporates ODE/SDE acceleration into their training objectives.

These models typically operate by making large jumps across time steps.
Consistency Models~\cite{cm,ict,ect,scm} formulate it as leaping from an intermediate time step directly to the end point of the trajectory.
Consistency Trajectory Models~\cite{ctm} aim to learn a trajectory between any two time steps, based on explicit integration (\ie ODE/SDE solving) during training.
Flow Map Matching~\cite{fmm} formulates the regression of the zeroth- and first-order derivatives of these flow fields.
Shortcut Models~\cite{shortcut} are built on the relationship between two time steps and their midpoint.
IMM~\cite{imm} leverages moment matching at different time steps.
MeanFlow~\cite{mf} formulates and parameterizes the average velocity across two arbitrary time steps.

Several improvements have been made to the MeanFlow formulation.
AlphaFlow \cite{alphaflow} decomposes the MeanFlow objective and adopts a schedule to interpolate from Flow Matching to MeanFlow.
Decoupled MeanFlow \cite{dmf} fine-tunes pre-trained Flow Matching models into MeanFlow by conditioning the final blocks of the networks on a second timestep.
CMT \cite{cmt} introduces mid-training using fixed explicit regression targets supplied by a pre-trained Flow Matching model before training the fastforward models. 
Our iMF is focused on the fundamental limitations of the MeanFlow objective, as well as the practical issue of CFG. These issues are orthogonal to other concurrent improvements.

%% file: sections/background.tex
\section{Background}
\label{sec:background}

\paragraph{Flow Matching.}
Flow Matching (FM)~\cite{fm,rectified,stochasticflow} learns a velocity field that flows between a prior distribution and the data distribution. 
We consider the standard linear schedule $z_t=(1-t)\,x+t\,e$ with data $x\!\sim p_{\text{data}}$ and noise $e\!\sim p_{\text{prior}}$ (\eg Gaussian). 
Computing the time-derivative gives a \emph{conditional} velocity $v_c=e-x$.
Flow Matching learns a network $v_\theta$ to regress $v_c$ by minimizing a loss function in the $v$-space (namely, $v$-loss):
\begin{equation}
\label{eq:fm-loss}
\mathbb{E}_{t,x,e}\big\|\,v_\theta(z_t,t)-(e-x)\,\big\|^2.
\end{equation}
As one $z_t$ can be given by multiple pairs of $(x,e)$, the underlying unique regression target is the \textit{marginal velocity}~\cite{fm}:
\begin{equation}
\label{eq:marginal-v} 
v(z_t,t) \triangleq \mathbb{E}[\,v_c\mid z_t\,], 
\end{equation}
which is marginalized over all pairs $(x,e)$ that satisfy $z_t$ at $t$.
For brevity, we omit $t$ and denote $v(z_t,t)$ as $v(z_t)$, and $v_\theta(z_t,t)$ as $v_\theta(z_t)$, in the remaining of this paper.

At generation time, FM samples by solving an ODE: $\ddt{z_t}\,=\,v_\theta(z_t)$.
This is done by a numerical solver (\eg Euler or Heun) integrating from $t=1$ to $0$, with $z_1\sim p_{\text{prior}}$.

\paragraph{MeanFlow.}
By viewing $v(z_t)$ as the \textit{instantaneous} velocity field, MeanFlow (MF)~\cite{mf} introduces the \emph{average} velocity field between two time steps $r$ and $t$:
\begin{equation}
\label{eq:avg-vel}
u(z_t, r, t)\; \triangleq \;\frac{1}{t-r}\int_{r}^{t} v(z_\tau)\,d\tau.
\end{equation}
Again, for brevity, we omit $r$ and $t$ and simply denote $u(z_t, r, t)$ by $u(z_t)$. 
Directly integrating \cref{eq:avg-vel} at training time is intractable.
Instead, MF takes the derivative \wrt $t$ and obtains a \textit{MeanFlow identity}~\cite{mf}:
\begin{equation}
\label{eq:mf-identity}
u(z_t)\;=\;v(z_t)\;-\;(t-r)\,\frac{d}{dt}\,u(z_t),
\end{equation}
which is used to establish a feasible training objective.
The term $\ddt u$ is given by~\cite{mf}:
\begin{equation}
    \label{eq:total-deriv}
    \ddt u(z_t) = \partial_z u(z_t)\,v(z_t) + \partial_t u(z_t)
    \triangleq \mathtt{JVP}(u; v) .
\end{equation}
This can be computed by Jacobian-vector product (JVP), between the Jacobian $[\partial_z{u}, \partial_r{u}, \partial_t{u}]$ and a tangent vector $[v, 0, 1]$.
Here, for brevity, we introduce the notation $\mathtt{JVP}(u; v)$ for this JVP computed at $u(z_t)$ and $v(z_t)$.

MeanFlow \textit{parameterizes} average velocity by a network $u_\theta(z_t)$ (conditioned on $r$ and $t$, omitted in notation for brevity). This network is optimized to approximate the MeanFlow identity (\ref{eq:mf-identity}).
In the formulation of the original MF, \cref{eq:mf-identity} is implemented as:
\begin{equation}
\label{eq:u-tgt}
    \utgt = (e-x) - (t-r)\,\mathtt{JVP}(u_\theta; e-x),
\end{equation}
Here, two approximations are made~\cite{mf}: (i) the marginal $v(z_t)$ is replaced with the conditional $v_c=e-x$, same as Flow Matching; (ii) the true $u$ in $\mathtt{JVP}$ is replaced by its network prediction $u_\theta$.
With this target $\utgt$, MF optimizes: 
\begin{equation}
\label{eq:mf-loss}
\mathbb{E}_{t,r,x,e}\,\|u_\theta- \text{sg}(\utgt)\|^2,
\end{equation}
where ``sg'' denotes stop-gradient, which helps create an \emph{apparent} target for training.
Once trained, MF directly performs one-step sampling via $z_0 = z_1 - u_\theta(z_1)$ given $(r,t)=(0,1)$, with $z_1\sim p_{\text{prior}}$.

%% file: sections/method.tex
\section{Improved Mean Flows}

\begin{figure}[t]
    \centering
    \vspace{-1em}
    \includegraphics[width=0.75\columnwidth]{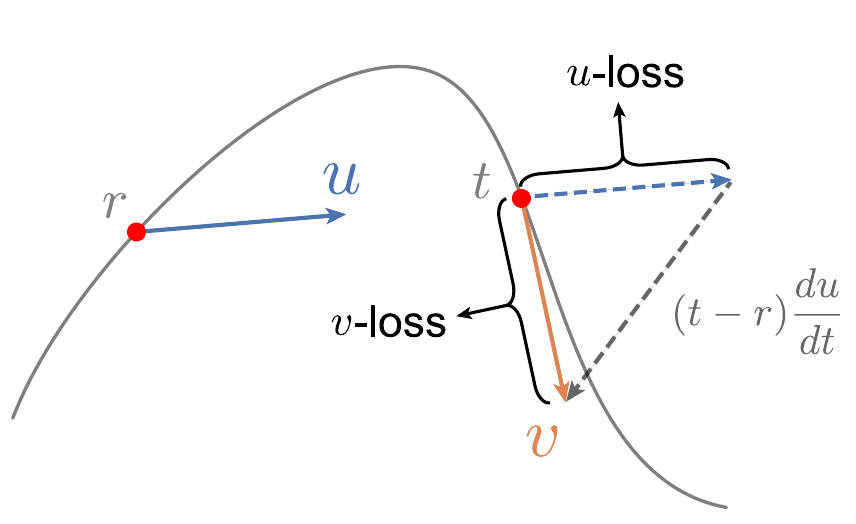}
    \caption{\textbf{MeanFlow as $v$-loss.}
    Original MeanFlow (MF)~\cite{mf} models the average velocity $u$ and train the network $u_\theta$ via a \textbf{$u$-loss} parameterized by $u_\theta$ itself. We show that MF can be reformulated as a \textbf{$v$-loss} re-parameterized by $u_\theta$, driven by the MeanFlow identity in \cref{eq:mf-identity2}.
    }
    \label{fig:imf-mf-teaser} 
\end{figure}

We identify and address two challenges of the original MF model. \textbf{(i)} The apparent target in \cref{eq:u-tgt} depends on the network. We aim for a more standard regression formulation (\cref{sec:v-pred}). \textbf{(ii)} To extend the MeanFlow identity for supporting CFG, original MF fixes the guidance scale before training. We relax this constraint and allow for flexible CFG (\cref{sec:method-cfg}). We then introduce an improved architecture for handling many types of conditions by in-context conditioning (\cref{sec:method-arch}).

\subsection{MeanFlow as \texorpdfstring{$v$-loss}{v-loss}}
\label{sec:v-pred}

\cref{eq:mf-loss} suggests that original MF~\cite{mf} is a \textbf{$u$-loss} parameterized by \textbf{$u$-pred}. In this subsection, we first show that the original MF can be reformulated as a \textbf{$v$-loss} (\ie instantaneous velocity) \textit{re-parameterized} by \textbf{$u$-pred}. This gives us a network-independent target. See \cref{fig:imf-mf-teaser}.

This reformulation reveals a hidden issue: the prediction function for $v$ requires access to \textit{unknown} quantities, not just $z_t$. We provide a solution to remedy this issue. With our reformulation, we arrive at a more standard regression problem. 

\paragraph{Reformulating MeanFlow as $v$-loss.}
While MF aims to compute $u$-loss, the true target $u$ is not accessible. As a result, the target $\utgt$ has a term approximated by $\mathtt{JVP}(u_\theta; e-x)$ in \cref{eq:u-tgt}, which is not a standard regression target. 
We notice that the \textit{instantaneous} velocity $v$ can serve as a more feasible target. We can rewrite the MeanFlow identity (\ref{eq:mf-identity}) as (see also \cref{fig:imf-mf-teaser}):
\begin{equation}
v(z_t) \;=\; u(z_t)\;+\;(t-r)\,\frac{d}{dt}\,u(z_t).
\label{eq:mf-identity2}
\end{equation}
Here, $v$ on the left-hand side can serve as a target, as in standard Flow Matching; the \textit{compound function} on the right-hand side can be parameterized by $u_\theta$. We denote the \mbox{(re-)parameterized} compound function as $V_\theta$:
\begin{equation}
    V_\theta
    \triangleq u_\theta\,(z_t) + (t-r)\,\sgjvp(u_\theta; e-x),
\label{eq:big-V}
\end{equation}
where $\sgjvp$ denotes stop-gradient on the $\ddt u_\theta$ outcome (we will discuss stop-gradient later).
Then we obtain a Flow Matching-like objective function, similar to \cref{eq:fm-loss}:
\begin{equation}
    \label{eq:V-loss}
    \mathbb{E}_{t,r,x,e}\|V_\theta - (e-x)\|^2.
\end{equation}
It is easy to show that the reformulation in (\ref{eq:big-V})(\ref{eq:V-loss}) is fully \textit{equivalent} to the original MF objective in (\ref{eq:u-tgt})(\ref{eq:mf-loss}).
This suggests that \textit{MeanFlow can be viewed as $v$-loss re-parameterized by $u_\theta$}. Such re-parameterization in \cref{eq:big-V} is driven by the MeanFlow identity in \cref{eq:mf-identity2}.

\begin{figure}[t]
    \centering
    \includegraphics[width=\columnwidth]{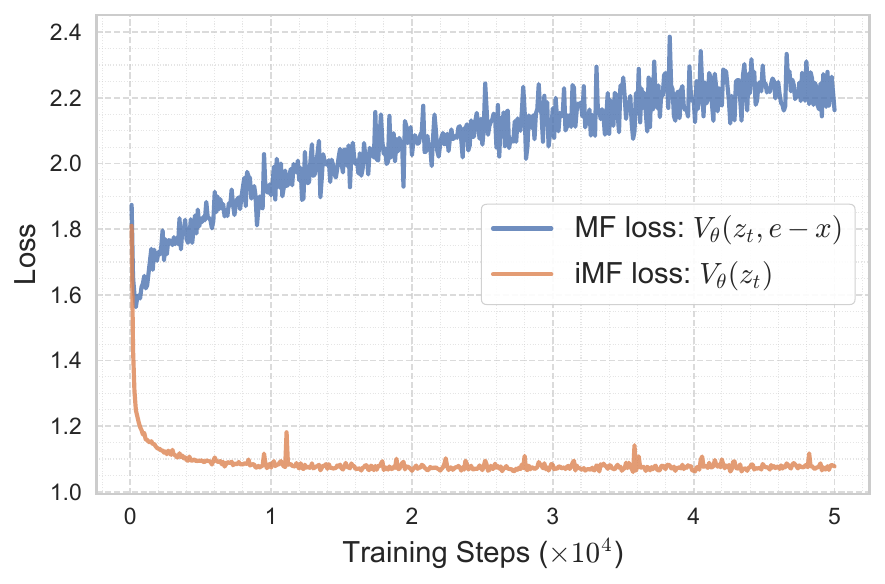}
    \vspace{-1.5em}
    \caption{\textbf{Training losses.}
    We examine the loss of samples only with $t \neq r$, since a batch also contains samples of $t = r$, for which the $\mathtt{JVP}$ term becomes zero due to its coefficient $(t-r)$.
    Both MF and iMF can be viewed as $v$-loss, using different forms of compound $V_\theta$.
    Original MF's loss is non-decreasing and has high variance.
    (Settings: MeanFlow-B/2, trained with basic $\ell_2$ loss with no adaptive weighting, and with no CFG.)
    } 
    \vspace{-.5em}
    \label{fig:loss} 
\end{figure}

This reformulation reveals a new issue: $V_\theta$ in \cref{eq:big-V} does not only take $z_t$ as input, but more importantly, also takes $e - x$ as another input. Formally, our parameterized compound function $V_\theta$ is:
\begin{equation}
V_\theta(z_t, e - x).
\end{equation}
This is illustrated in \cref{fig:teaser}\textbf{(a)}.
From the perspective of a standard regression formulation (\eg \cref{eq:fm-loss}), this is not a fully legitimate prediction function. We will show the negative effect of this extra input in \cref{fig:loss}. 

\input{sections/code}

\paragraph{Improved MeanFlow Parameterization.}
The reason for $V_\theta$'s dependence on $e-x$ is on $\mathtt{JVP}$, which can be traced back to the approximation in \cref{eq:u-tgt}: the \textit{marginal} velocity $v$ in \cref{eq:total-deriv} is replaced by the \textit{conditional} velocity $v_c=e-x$. Rather than doing this replacement, we can parameterize the marginal $v$ instead. Formally, we re-define the compound function $V_\theta$ as:
\begin{equation}
    V_\theta\,(z_t) \\
    \triangleq u_\theta\,(z_t) + (t-r)\,\sgjvp\left(u_\theta; \textcolor[RGB]{255, 64, 64}{v_\theta}\right).
\label{eq:imf-vpred}
\end{equation}
Here, inside the function of $\mathtt{JVP}$, both $u_\theta$ and $v_\theta$ are network predictions: both take $z_t$ as the sole input. As such, our $V_\theta$ takes only $z_t$ as the input, which is thus a legitimate prediction function. See \cref{fig:teaser}\textbf{(b)}.

To realize ${v_\theta}$ with minimal overhead, we can reuse all or most of the network $u_\theta$.
We propose two solutions:
\begin{itemize}[leftmargin=1em, rightmargin=0em]
    \item \textbf{Boundary condition} of $u_\theta$. By definition, we have the relation: $v(z_t, t) \equiv u(z_t, t, t)$, that is, $v$ equals $u$ at $r\rightarrow t$. As such, we can simply represent $v_\theta(z_t, t)$ by $u_\theta(z_t, t, t)$. This solution introduces no extra parameters. We empirically show that this is sufficient for addressing the issue we consider here.
    \item \textbf{Auxiliary $v$-head.} Beyond directly reusing $u_\theta(z_t, t, t)$, we can add an auxiliary head in the network of $u_\theta(z_t, r, t)$ that serves as a subnetwork $v_\theta$. This introduces extra capacity for modeling $v$. It is at the cost of extra \textit{training-time} parameters, which, however, are not used at inference-time (as only $u_\theta$ is needed). 
    More details are in appendix.
    Using this head improves the results further.
\end{itemize}

\noindent
The pseudo-code of our iMF formulation is in \cref{alg:code}, where the simpler form of $v_\theta(z_t, t) \equiv u_\theta(z_t, t, t)$ is shown.

\paragraph{Comparison and Analysis.} 
In \cref{fig:loss}, we compare the training loss between the original MF and the iMF objectives (\textit{without} auxiliary $v$-head). We examine only the samples with $t \neq r$, as the $\mathtt{JVP}$ term becomes zero when $t = r$ and thus is not our focus.

Although the two formulations only differ in $V_\theta(z_t, e-x)$ and $V_\theta(z_t)$, this distinction results in strikingly different behavior. The original MF's loss has a much \textit{higher variance} and is non-decreasing\footnotemark, even though its objective can still successfully enable one-step generation.

\footnotetext{If we also include the samples of $t = r$, the original MF's overall loss can still decrease, depending on the portion of such samples.}

This comparison may look counterintuitive, because the form of $V_\theta(z_t, e-x)$ seems to ``leak'' the regression target. However, we note that the true, unique regression target for $v$-loss is not the conditional velocity $e-x$, but the marginal $v(z_t) = \mathbb{E}[\,v_c\mid z_t\,]$ in \cref{eq:marginal-v}, and therefore the leaking does not directly disclose the true $v(z_t)$. On the other hand, according to \cref{eq:total-deriv}, the input to $\mathtt{JVP}$ should not be the conditional $v_c = e-x$, but should be the marginal $v(z_t)$. 
As this is the input tangent vector to $\mathtt{JVP}$, the variance of the conditional velocity can be significantly magnified by $\mathtt{JVP}$ (\ie the Jacobian-\emph{vector} product). 
Our tangent vector is predicted by $v_\theta(z_t)$, which should have lower variance than $e - x$.
\cref{fig:loss} suggests that the large variance dominates the resulting loss.

\paragraph{About Stop-gradient.} Our formulation does not remove the stop-gradient operation, as indicated by the notation $\sgjvp$. Unlike original MF, in our case the stop-gradient is part of the prediction function $V_\theta$, not the regression target. As such, in principle, this stop-gradient is not strictly needed for the formulation itself. However, in practice, we observe that using the stop-gradient inside $V_\theta$ is still beneficial, as removing it introduces high-order gradients \wrt $\theta$ and makes optimization more difficult. 

\subsection{Flexible Guidance}
\label{sec:method-cfg}

\begin{figure}[t]
\centering
\includegraphics[width=\linewidth]{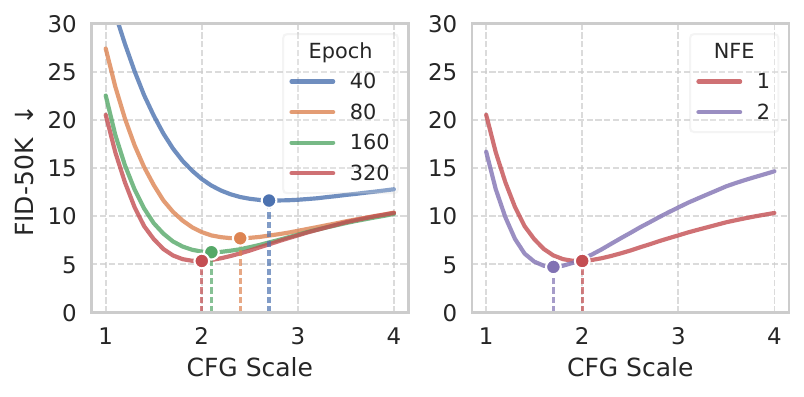}
\vspace{-1.5em}
\caption{
\textbf{Optimal CFG scales shift under different settings}.
In general, a \textit{stronger} setting has a \textit{smaller} optimal CFG scale, as reflected by increased training epochs (left) and inference steps (right).
This investigation is enabled by our flexible CFG-conditioning, where a single model can support varying CFG scales even in the single/few-NFE case. (Settings: iMF-B/2 on ImageNet 256$\times$256.)
}
\label{fig:train-inf-cfg}
\end{figure}

Thus far, we have not discussed the formulation of classifier-free guidance (CFG)~\cite{cfg}. 
The original MF~\cite{mf} proposed a formulation to support 1-NFE CFG, provided that a guidance scale is \textit{fixed} at training-time.
However, a fixed guidance scale sacrifices the flexibility of adjusting this core hyperparameter at inference time. More importantly, the optimal CFG scales \textit{shift} under different settings (\cref{fig:train-inf-cfg}), and in general, a \textit{strong} model (\eg larger size, longer training, and/or more NFEs) favors a \textit{smaller} CFG scale. It is suboptimal to freeze the scale \textit{a priori}.

To address this issue, we reformulate the CFG scale as a form of \textit{conditioning}, analogous to how a model is conditioned on time steps (\eg $t$ and $r$). This enables the scale to vary at training and inference time.

\paragraph{Original MeanFlow with fixed guidance}. 
The original MF \cite{mf} considers a \textit{fixed} guidance field $v_{\text{cfg}}$: 
\begin{equation}
    \label{eq:cfg}
       v_{\text{cfg}}(z_t\mid\mathbf{c}) = \omega\,v(z_t\mid\mathbf{c}) + (1-\omega)\,v(z_t),
\end{equation}
where $\mathbf{c}$ is the class-condition, and $\omega$ is a fixed guidance scale.
Combining this definition and the MeanFlow identity, the original MF learns a \textit{class-conditional} average velocity, namely, $u_{\theta}(z_t\mid\mathbf{c})$. We omit the derivations here and refer readers to~\cite{mf}; but analogous to our reformulation in \cref{sec:v-pred}, conceptually, we can re-parameterize $u_{\theta}(z_t\mid\mathbf{c})$ into a compound function: 

\begin{equation}
    V_{\theta}(\cdot \mid \mathbf{c})
    \triangleq u_{\theta}(z_t\mid\mathbf{c}) + (t-r)\,\sgjvp.
    \label{eq:bigV-cfg}
\end{equation}
Here, for brevity, we omit the input to $V_{\theta}$ (and to $\mathtt{JVP}$), which is not our focus in this subsection: our discussion here can support both objectives in \cref{eq:big-V} (original MF) and \cref{eq:imf-vpred} (iMF).
This $V_{\theta}$ is trained to fit a target determined by a fixed $\omega$ in the original MF \cite{mf}.

\paragraph{Improved MeanFlow with flexible guidance}. 
If the underlying guidance field  $v_{\text{cfg}}$ (\ref{eq:cfg}) is given by different $\omega$ values, we can still let our neural network fit each of them. To do so, we only need to allow the network to \textit{condition} on the CFG guidance scale $\omega$.
Similar strategies have been studied in multi-step methods \cite{guidancedist,nocfg,modelguidance}, which we extend to our one-step method here.

Formally, we extend \cref{eq:bigV-cfg} to:
\begin{equation}
    V_{\theta}(\cdot \mid \mathbf{c},\,\highlight{\omega})
    \triangleq u_{\theta}(z_t\mid\mathbf{c},\,\highlight{\omega}) + (t-r)\,\sgjvp,
\end{equation}
which indicates that our compound function $V_{\theta}$ can be conditioned on $\omega$, and this conditioning is handled by the network $u_{\theta}$. This is analogous to standard time-conditioning (\eg $t$ and $r$), which turns a continuous value into a learnable embedding.
At training time, the value of $\omega$ is randomly sampled from a given distribution. 
The implementation details of training with CFG conditioning is in appendix. 

\cref{fig:train-inf-cfg} shows the effect of our flexible CFG in iMF. Under different training and inference settings, the optimal guidance scale varies. Even for the \textit{same} model, training longer or using more inference steps can favor a different guidance scale, and therefore it is impossible to find the optimal scale beforehand. Our design unlocks the full potential of CFG for 1-NFE models.

\paragraph{Additional guidance conditioning.} Our formulation not only enables conditioning on a single variable $\omega$, but also allows for other guidance-related factors. We can handle CFG interval \cite{interval} under the same paradigm.

CFG interval \cite{interval} is an effective technique for improving sample diversity. In its original definition, it applies CFG only to a time interval $[t_\textrm{min}, t_\textrm{max}]$ at \textit{inference-time}. To support this behavior at training time in our one-step model, we can also view the two values $t_\textrm{min}$, $t_\textrm{max}$ as a form of conditioning. At training time, when $t$ is outside of this interval, CFG is disabled (by setting $\omega=1$).

We use the notation $\Omega=\{\omega, t_\textrm{min}, t_\textrm{max}\}$ to denote all conditions related to CFG. Each item in $\Omega$ has its own embedding. All conditions can be handled by standard adaLN-zero, or in-context conditioning, discussed next.

\begin{figure}[t]
    \centering
    \vspace{-1em}
    \includegraphics[width=0.8\columnwidth, angle=0,trim={0cm 0cm 0cm 0cm}, clip]{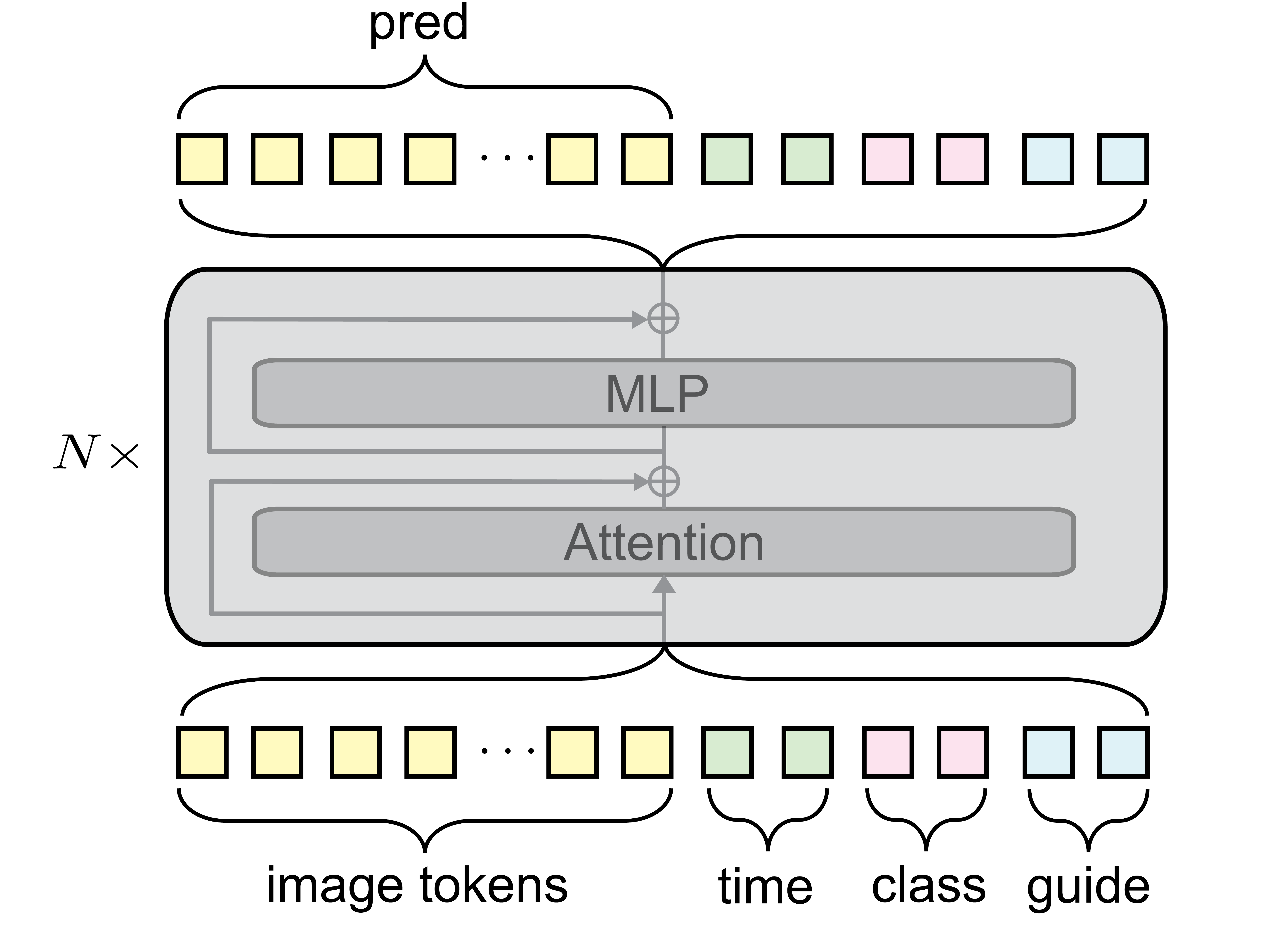}
    \vspace{-1em}
    \caption{\textbf{Improved in-context conditioning.}
    Each type of conditions is turned into \textit{multiple} tokens, which are concatenated with the image latent tokens along the sequence axis. It accommodates the conditions of time steps $(r,t)$, class $\mathbf{c}$, and guidance-related factors $\Omega$ (CFG scale $\omega$ and CFG intervals). Importantly, \textbf{\textit{we do not use adaLN-zero for conditioning}}, which significantly reduces the model size (number of parameters) while maintaining performance.
    }
    \label{fig:network} 
    \vspace{-1em}
\end{figure}

\subsection{Improved In-context Conditioning}
\label{sec:method-arch}

Our model has \textit{a diverse set} of conditions, including two time steps $r$ and $t$, a class label $\mathbf{c}$, and the guidance-related conditions $\Omega$. In its complete notation, the network $u_\theta$ is:
\begin{equation}
\label{eq:cond-args}
u_\theta =u_\theta\,\left(z_t \mid r,\,t,\,\mathbf{c},\,\Omega \right).
\end{equation}
Typically, the conditioning is handled by adaLN-zero \cite{dit}, which \textit{sums} all condition embeddings. When many heterogeneous conditions are present, summing their embeddings and processing by adaLN-zero may become less effective, as this single operation can be overburdened.

\paragraph{Improved MeanFlow conditioning.}
To handle these many conditions, we resort to the \textit{in-context} conditioning strategy. In-context conditioning was explored in DiT \cite{dit} but was found inferior to adaLN-zero in their setting. We find that this gap can be closed if \textit{multiple} tokens are used for each condition. In our implementation, we use 8 tokens for class, and 4 tokens for each other conditions (see appendix). 
All these learnable tokens are concatenated along the sequence axis, jointly with the tokens from images (in the latent space, same as DiT \cite{dit}). The sequence is processed by Transformer blocks (\cref{fig:network}). 
This architecture enables us to accommodate different types of conditions flexibly.

As an important by-product, our in-context conditioning enables us to completely \textit{remove} adaLN-zero, which is parameter-heavy. This yields a \textbf{1/3 reduction} in model size (\eg from 133M to 89M for our iMF-Base model) when depth and width are unchanged. This also allows us to design the larger models more flexibly.

%% file: sections/code.tex
\definecolor{codeblue}{rgb}{0.25,0.5,0.5}
\definecolor{codesign}{RGB}{0, 0, 255}
\definecolor{codefunc}{rgb}{0.85, 0.18, 0.50}

\lstdefinelanguage{PythonFuncColor}{
  language=Python,
  keywordstyle=\color{blue}\bfseries,
  commentstyle=\color{codeblue},
  stringstyle=\color{orange},
  showstringspaces=false,
  basicstyle=\ttfamily\small,
  literate=
    {+}{{\color{codesign}+ }}{1}
    {-}{{\color{codesign}- }}{1}
    {*}{{\color{codesign}* }}{1}
    {/}{{\color{codesign}/ }}{1}
    {sample_t_r}{{\color{codefunc}sample\_t\_r}}{1}
    {randn_like}{{\color{codefunc}randn\_like}}{1}
    {jvp}{{\color{codefunc}jvp}}{1}
    {stopgrad}{{\color{codefunc}stopgrad}}{1}
    {metric}{{\color{codefunc}metric}}{1}
}

\lstset{
  language=PythonFuncColor,
  backgroundcolor=\color{white},
  basicstyle=\fontsize{9pt}{9.9pt}\ttfamily\selectfont,
  columns=fullflexible,
  breaklines=true,
  captionpos=b,
}

\newcommand{\safeColorbox}[2]{%
  \begingroup
  \setlength{\fboxsep}{0pt}%
  \colorbox{#1}{\strut #2}%
  \endgroup
}

\begin{algorithm}[t]
\centering
\caption{{improved MeanFlow}: training.\\
{\scriptsize Note: in PyTorch and JAX, \texttt{jvp} returns the function output and JVP.}}
\label{alg:code}
\begin{minipage}{0.98\linewidth}
\begin{lstlisting}[language=PythonFuncColor, escapechar=`]
# fn(z, r, t): function to predict u
# x: training batch

t, r = sample_t_r()
e = randn_like(x)

z  = (1 - t) * x + t * e

# instantaneous velocity v at time t
v = fn(z, t, t)

# predict u and dudt
u, dudt = jvp(fn, (z, r, t), (v, 0, 1))

# compound function V
V = u + (t - r) * stopgrad(dudt)
error = V - (e - x)

loss = metric(error)
\end{lstlisting}
\end{minipage}
\end{algorithm}

%% file: sections/experiments.tex
\section{Experiments}

Our experiment settings follow those of the original MeanFlow \cite{mf}, using the same public code.
The experiments are on ImageNet \citep{imagenet} class-conditional generation at 256${\times}$256 resolution. 
Following \cite{shortcut,imm,mf}, the model operates on the latent space of a pretrained VAE tokenizer~\citep{ldm}, which produces 32${\times}$32${\times}$4 latents from 256${\times}$256${\times}$3 images.

We evaluate the challenging protocol of \textbf{1-NFE} generation, where all our models are trained \textbf{\emph{from scratch}}.
We report Fr\'echet Inception Distance (FID)~\citep{fid} on 50K generated images (with additional metrics in \cref{tab:extra-metrics}).
Detailed configurations are in appendix.

\paragraph{Baseline.} 
In our ablations, we use the MeanFlow-B/2 model \cite{mf}.
The ablation models are trained for 240 epochs. Our setting is exactly the same as that of MeanFlow-B/2 in \cite{mf}, which has a 1-NFE FID of 6.17 (with CFG). This model is our starting point.

\begin{figure}[t]
\vspace{-1em}
\centering
\includegraphics[width=.9\linewidth]{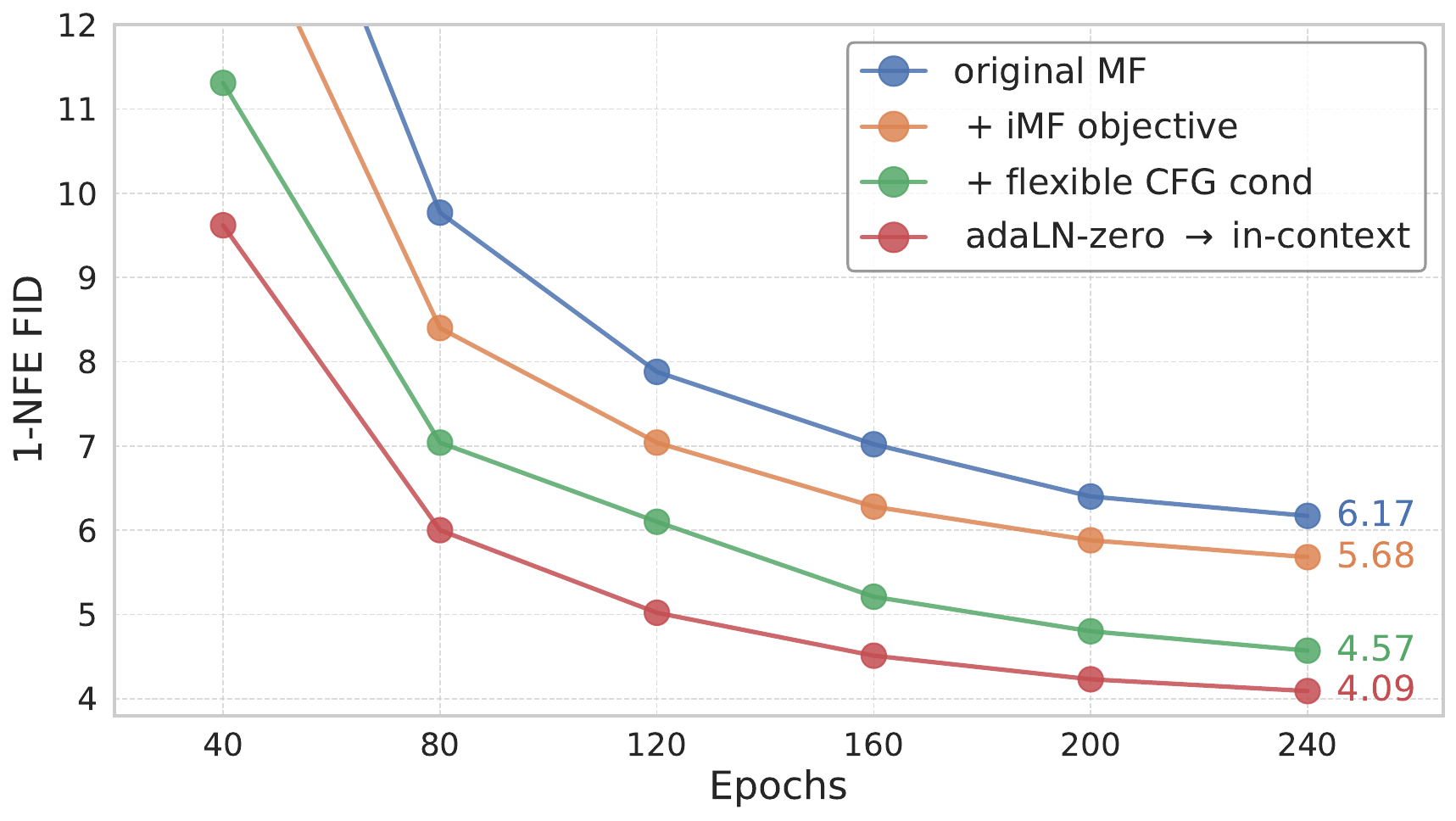}
\vspace{-0.7em}
\caption{
\textbf{FID curves during training.} 
The original MeanFlow-B/2 baseline has a 1-NFE FID of 6.17. Using the improved training objective (\cref{sec:v-pred}), FID improves to 5.68. Incorporating flexible CFG conditioning (\cref{sec:method-cfg}) reduces FID to 4.57. Replacing adaLN-zero with in-context conditioning (\cref{sec:method-arch}) further improves FID to 4.09. See also \cref{tab:fid_ablation}.
}
\vspace{-1em}
\label{fig:learning-ablation}
\end{figure}

\subsection{Ablation Study}
\label{sec:ablation}

In \cref{tab:fid_ablation}\textbf{(a)} and \textbf{(b)}, we ablate the iMF designs discussed in \cref{sec:v-pred} and \cref{sec:method-cfg}. The architectural improvements are in \cref{tab:fid_ablation}(c). FID curves during training are in \cref{fig:learning-ablation}.

\paragraph{MeanFlow as $v$-loss.}
In \cref{tab:fid_ablation}\textbf{(a)}, we compare the original MF training formulation (\ref{eq:mf-loss}) with our iMF training formulation (\ref{eq:imf-vpred}). 
We do not use CFG-conditioning here.
We compare two variants of computing $v_\theta$ for the $\mathtt{JVP}$ usage in (\ref{eq:imf-vpred}): the boundary condition or auxiliary head.

When using the variant of boundary condition  ($v_\theta = u_\theta(z_t,t,t)$), our formulation improves the case w/o CFG from an FID of 32.69 to 29.42, representing a solid gain of 3.27. This variant adds no extra parameters at training or inference time. This result demonstrates the impact of the legitimate regression formulation.

When using the auxiliary head variant, our formulation also substantially improves over the original MF, from 32.69 to 30.76 w/o CFG, with a gain of 1.93. While the gain is smaller than that of using the boundary condition, it becomes relatively more significant in the case of ``w/ CFG'', suggesting a more capable model is desired to handle the more challenging scenario.

\input{tables/ablation}

In the case of ``w/ CFG'' (here, trained with a fixed $\omega$), using the boundary condition improves FID from 6.17 to 5.97 (\cref{tab:fid_ablation}\textbf{(a)}). While this relative gain is smaller, we observe that the same setting has a more pronounced impact on the same MF-XL model:
\begin{center}
\vspace{-.3em}
\centering\footnotesize
\begin{tabular}{y{120}|c}
\textbf{FID, 1-NFE} & \textbf{MF-XL/2 model}, w/ CFG \\
\hline
original MF~\cite{mf}     
    & 3.43 \\
our $V_\theta$, with $v_\theta = u_\theta(z_t,t,t)$       
    & \textbf{2.99} \\
\end{tabular}
\vspace{-.3em}
\end{center}
We hypothesize that when the model has more capacity, it can better leverage the capacity to learn $v_\theta$ by $u_\theta(z_t,t,t)$, and therefore benefits more from this formulation.

Further, \cref{tab:fid_ablation}\textbf{(a)} shows that the auxiliary head achieves an FID of 5.68 w/ CFG, which is about 10\% relative improvement over the original MF.
This auxiliary head introduces no extra parameters or compute at inference time.
All these comparisons demonstrate that a reliable $v$ estimation \textit{as $\mathtt{JVP}$'s input} is critical for MeanFlow methods.

\paragraph{Flexible guidance.}
In \cref{tab:fid_ablation}\textbf{(b)}, we examine the CFG conditioning proposed in \cref{sec:method-cfg}. This ablation is best examined together with \cref{fig:train-inf-cfg}: the major advantage of CFG conditioning is on the inference-time flexibility, which may not be simply reflected by a single FID number.

In \cref{tab:fid_ablation}\textbf{(b)}, when using the simpler $\omega$-conditioning (\ie only on the CFG scale $\omega$), the FID w/ CFG improves slightly from 5.68 to 5.52. This marginal gain is unsurprising, because the original MF \cite{mf} already had a near-optimal but fixed training-time  $\omega$, for this small model. This gain is more substantial for larger models, for which searching for a fixed training-time $\omega$ becomes impractical.

\cref{tab:fid_ablation}\textbf{(b)} further shows that richer CFG-conditioning (\ie on $\Omega$) substantially improves the FID, by 1.11 to 4.57. This gain is because $\Omega$-conditioning enables CFG interval \cite{interval} \textit{at inference time}, and CFG interval is highly effective even for multi-step methods. 
Our conditioning strategy does not affect the 1-NFE sampling behavior: $(t_\textrm{min}, t_\textrm{max})$ are turned into embeddings for 1-NFE generation.

Interestingly, our CFG conditioning also enables us to mimic the ``w/o CFG'' behavior at inference time. We achieve this by setting $\omega=1.0$ at inference, which represents the ``no CFG'' case (see \cref{eq:cfg}). \cref{tab:fid_ablation}\textbf{(b)} shows that the FID at $\omega=1.0$ (``w/o CFG'') is significantly improved by 10 points, from 30.76 to 20.95. This suggests that training our models across a range of CFG scales can improve their \textit{generalization} performance, substantially improving their results even at a suboptimal $\omega$ value.

\input{tables/v1_v2}

\input{tables/imgnet-256}

\paragraph{In-context conditioning.} 
Thus far, our ablations have been using adaLN-zero \cite{dit} for conditioning. 
In \cref{tab:fid_ablation}\textbf{(c)}, we replace it with our multi-token in-context conditioning (\cref{sec:method-arch}). As adaLN-zero is parameter-heavy, removing it yields a substantial 1/3 reduction in model size, from 133M to 89M. 
Such a reduction is highly attractive for larger models.
The FID is improved from 4.57 to 4.09.

Finally, following \cite{vavae}, we incorporate general-purpose Transformer improvements: SwiGLU \cite{swiglu}, RMSnorm \cite{rmsnorm}, and RoPE \cite{rope}. These components put together improve FID from 4.09 to 3.82.  Training longer yields an extra gain, achieving 3.39 FID with this B-size model.

\subsection{Comparisons with Original MeanFlow} 

In \cref{tab:extra-metrics}, we provide a system-level comparison with the original MF \cite{mf}. We note that removing adaLN-zero makes it impossible to fully calibrate the model sizes, and as such, the B/M/L/XL notations are mainly for the ease of referring. In our designs: \textbf{(i)} the B-size model has the same depth and width in both MF and iMF; \textbf{(ii)} the M-size model is designed to have \textit{smaller} size and \textit{less} compute; and \textbf{(iii)} the L/XL-size models are designed to roughly match the model size of the MF counterparts (yet are still $\app$10\% smaller).

Overall, \cref{tab:extra-metrics} shows that our iMF models have substantially better FID and IS results.
Our iMF-XL/2 model achieves a 1-NFE FID of \textbf{1.72}, representing a \textbf{50\%} relative reduction compared to MF-XL/2’s 3.43. Qualitative examples are in \cref{fig:result_images} and appendix.

\input{tables/visuals}

\subsection{Comparisons with Previous Methods}  

In \cref{tab:all_results}, we provide system-level comparisons with previous methods. We categorize the methods into:
(i) fastforward generative models trained \textit{from scratch} (\cref{tab:all_results}, left); 
(ii) fastforward generative models, \textit{distilled} from pre-trained multi-step models (\cref{tab:all_results}, mid-top); 
(iii) multi-step diffusion/flow models (\cref{tab:all_results}, mid-bottom); (iv) GAN and autoregressive models (\cref{tab:all_results}, right). 

\paragraph{Fastforward models from scratch.} \cref{tab:all_results} (left) shows that our iMF substantially outperforms other fastforward models that are also trained from scratch. 
In addition, our 1-NFE FID of 1.72 also outperforms those \textit{distilled} from pre-trained models (\cref{tab:all_results}, mid-top), suggesting that training from scratch can produce highly competitive fastforward models.
When relaxing NFE to 2, iMF achieves an FID of \textbf{1.54}. This further closes the gap with the \textit{many-step} diffusion/flow models (\cref{tab:all_results}, mid-bottom).

%% file: tables/ablation.tex
\begin{table}[t]
\centering
\begin{minipage}{\linewidth}
\begin{subtable}{\linewidth}
\centering
\tablestyle{4pt}{1.1}
\begin{tabular}{y{120}|x{45}|x{45}}
 \textbf{FID, 1-NFE} & w/o CFG & w/ CFG  \\
\hline
original MF~\cite{mf}     & 32.69 & 6.17 \\
our $V_\theta$, with $v_\theta =u_\theta(z_t,t,t)$       & \textbf{29.42} & 5.97 \\
our $V_\theta$, with $v_\theta$ from aux. head         & 30.76 & \textbf{5.68} \\
\end{tabular}
\caption{\textbf{MeanFlow as $v$-loss}. We compare with original MF with our iMF objective in \cref{eq:imf-vpred} in \cref{sec:v-pred}.
We compare two variants of computing $v_\theta$ for \cref{eq:imf-vpred}, namely, using $u_\theta$'s boundary condition or an auxiliary head.
In each row, ``w/o CFG'' and ``w/ CFG'' are two models trained separately, as is in original MF (which does not support flexible inference-time CFG). 
}
\label{subtab:v-pred}
\end{subtable}
\vspace{-0.25em}
\\
\begin{subtable}{\linewidth}
\centering
\tablestyle{4pt}{1.1}
\begin{tabular}{y{120}|x{45}|x{45}}
 \textbf{FID, 1-NFE} & w/o CFG & w/ CFG  \\
\hline
best in \cref{tab:fid_ablation}\textbf{(a)}        &  30.76 & 5.68 \\
CFG-condition: $\omega$-condition    &  25.15 & 5.52 \\
CFG-condition: $\Omega$-condition &  \textbf{20.95} & \textbf{4.57} \\
\end{tabular}
\caption{\textbf{Flexible guidance}. Adding guidance as conditioning enables flexible guidance at inference (\cref{sec:method-cfg}).
$\omega$-condition is the basic conditioning on the guidance scale $\omega$.
$\Omega$-condition allows to further condition on CFG interval's start and end points.
Here, only in the first row, ``w/o CFG'' and ``w/ CFG'' are two models trained separately, same as \cref{tab:fid_ablation}\textbf{(a)}; when using our flexible CFG (last two rows), the ``w/o CFG'' cases are simply $\omega=1$ at inference-time, using a single trained model.
}
\label{subtab:flexible-cfg}
\end{subtable}
\vspace{-0.25em}
\\
\begin{subtable}{\linewidth}
\centering
\tablestyle{4pt}{1.1}
\begin{tabular}{y{120}|x{45}|x{45}}
\textbf{FID, 1-NFE} & \# params & w/ CFG \\
\hline
best in \cref{tab:fid_ablation}\textbf{(b)}               & 133M & 4.57             \\
{adaLN-zero $\rightarrow$ in-context cond.} & \textbf{89M} & \textbf{4.09}              \\
\hline
 \deemph{$+$ advanced Transformer blocks} & \deemph{89M} & \deemph{3.82}              \\
\deemph{$+$ longer training (640ep)}  & 
\deemph{89M} & \deemph{3.39} \\
\end{tabular}
\caption{\textbf{In-context conditioning} and other improvements.
Replacing adaLN-zero \cite{dit} with our multi-token in-context conditioning (\cref{sec:method-arch}) improves the results and substantially reduces the model size. Advanced Transformer blocks and longer training yield improvements as expected.}
\label{subtab:arch}
\end{subtable}
\end{minipage}
\vspace{-0.5em}
\caption{\textbf{Ablation study on 1-NFE generation.} FID-50K is evaluated on ImageNet 256$\times$256. All are with the MF-B/2 backbone, trained for 240 epochs from scratch by default.
}
\label{tab:fid_ablation}
\end{table}

%% file: tables/v1_v2.tex
\begin{table}[t]
\small
\centering
\tablestyle{4pt}{1.1}
\resizebox{\linewidth}{!}{
\begin{tabular}{y{40}|x{20}x{20}|x{30}x{24}|x{24}x{24}}
{config}       & depth & width & {\# params} & Gflops & {FID}$\downarrow$ & {IS}$\uparrow$ \\
\hline
{MF}-B/2       & 12 & 768  & 131M & 23.1  & 6.17 & 208.0   \\
{MF}-M/2       & 16 & 1024 & 308M & 54.0  & 5.01 & 252.0   \\
{MF}-L/2       & 24 & 1024 & 459M & 80.9  & 3.84 & 250.9   \\
{MF}-XL/2      & 28 & 1152 & 676M & 119.0 & 3.43 & 247.5   \\
\hline
\textbf{iMF}-B/2       & 12 & 768 & 89M  & 24.9  & 3.39 & 255.3    \\
\textbf{iMF}-M/2       & 24 & 768 & 174M & 49.9  & 2.27 & 257.7    \\
\textbf{iMF}-L/2       & 32 & 1024 & 409M & 116.4 & 1.86 & 276.6   \\
\textbf{iMF}-XL/2      & 48 & 1024 & 610M & 174.6 & \textbf{1.72} & \textbf{282.0}   \\
\end{tabular}
}
\vspace{-.75em}
\caption{\textbf{System-level comparison with original MeanFlow}, evaluated by FID and IS \cite{improvedgan} on ImageNet 256$\times$256 with \textbf{1-NFE} generation.
The notations of B/M/L/XL are mainly for reference, as it is impossible to calibrate both model size (\# params) and compute (Gflops) due to the removal of adaLN-zero.
The compute is for 1-NFE of the generator, excluding the tokenizer decoder.
}
\label{tab:extra-metrics}
\vspace{-1.35em}
\end{table}

%% file: tables/imgnet-256.tex
\begin{table*}[t]
\centering
\vspace{-1.5em}
\resizebox{!}{0.15\linewidth}{
\setlength{\tabcolsep}{4pt}
\small
\begin{tabular}{y{88}x{34}x{24}x{24}}
\toprule
{Method} & {\# Params} & NFE & {FID} \\
\midrule
\rowcolor[gray]{0.9}
\multicolumn{4}{l}{\textit{\textbf{1-NFE diffusion/flow from scratch}}} \\
~~iCT-XL/2~\citep{ict}               & 675M & 1 & 34.24   \\
~~Shortcut-XL/2~\citep{shortcut} & 675M & 1 & 10.60 \\
~~MeanFlow-XL/2~\citep{mf} & 676M & 1 & 3.43   \\
~~TiM-XL/2~\citep{tim} & 664M & 1 & 3.26 \\
~~$\alpha$-Flow-XL/2+~\citep{alphaflow} & 676M & 1 & 2.58 \\
\midrule
~~\textbf{iMF}-B/2 (ours)      & 89M  & 1 & 3.39             \\
~~\textbf{iMF}-M/2 (ours)      & 174M & 1 & 2.27             \\
~~\textbf{iMF}-L/2 (ours)      & 409M & 1 & 1.86             \\
~~\textbf{iMF}-XL/2 (ours)     & 610M & 1 & \textbf{1.72}    \\
\midrule
\rowcolor[gray]{0.9}
\multicolumn{4}{l}{\textit{\textbf{2-NFE diffusion/flow from scratch}}} \\
~~iCT-XL/2~\citep{ict}              & 675M & 2 & 20.30   \\
~~IMM-XL/2~\citep{imm}              & 675M & 1$\times$2 & 7.77 \\
~~MeanFlow-XL/2+~\citep{mf}                  & 676M & 2 & 2.20  \\
~~$\alpha$-Flow-XL/2+~\citep{alphaflow} & 676M & 2 & 1.95\\
\midrule
~~\textbf{iMF}-XL/2 (ours)                       & 610M & 2 & \textbf{1.54}   \\
\bottomrule
\end{tabular}
}
\hspace{.1em}
\resizebox{!}{0.15\linewidth}{
\setlength{\tabcolsep}{4pt}
\textcolor[rgb]{0.6,0.6,0.6}{
\small
\begin{tabular}{y{92}x{34}x{30}x{18}}
\toprule
{Method} & {\# Params} & NFE & FID \\
\midrule
\multicolumn{4}{l} {\textit{\textbf{1-NFE diffusion/flow (distillation)}}} \\
~~$\pi$-Flow-XL/2~\citep{piflow} & 675M & 1 & 2.85\\
~~{DMF-XL/2+}~\cite{dmf} & 675M &  1 & 2.16\\
~~FACM-XL/2~\cite{facm} & 675M & 1 & 1.76 \\
\midrule
\multicolumn{4}{l} {\textit{\textbf{Multi-NFE diffusion/flow}}} \\
~~ADM-G~\cite{adm}  & 554M & 250$\times$2 & 4.59 \\
~~LDM-4-G~\cite{ldm}      & 400M & 250$\times$2 & 3.60 \\
~~SimDiff~\cite{simdiff}  & 2B & 1000$\times$2 & 2.77 \\
~~DiT-XL/2~\cite{dit} & 675M & 250$\times$2 & 2.27\\
~~SiT-XL/2~\cite{sit} & 675M   & 250$\times$2 & 2.06  \\
~~SiT-XL/2\;+\;REPA~\cite{repa}      & 675M & 250$\times$2 & 1.42 \\
~~SiD2~\cite{sid2}   & -- & 512$\times$2 & 1.38 \\
~~LightningDiT-XL/2~\cite{vavae} & 675M & 250$\times$2 & 1.35 \\
~~DDT-XL/2~\cite{ddt} & 675M  & 250$\times$2 & 1.26 \\
~~RAE~\cite{rae}\;+\;DiT$^\text{DH}$-XL & 839M & 50$\times$2 & 1.13 \\
\bottomrule
\end{tabular}
}
}
\hspace{.1em}
\resizebox{!}{0.15\linewidth}{
\setlength{\tabcolsep}{4pt}
\textcolor[rgb]{0.6,0.6,0.6}{
\small
\begin{tabular}{y{80}x{34}x{30}x{20}}
\toprule
{Method} & {\# Params} & {NFE} & FID \\
\midrule
\multicolumn{4}{l}{\textit{\textbf{GANs}}} \\
~~BigGAN~\citep{biggan}          & 112M & 1 & 6.95 \\
~~GigaGAN~\citep{gigagan}        & 569M & 1 & 3.45 \\
~~StyleGAN-XL~\citep{styleganxl} & 166M & 1 & 2.30 \\
\midrule
\multicolumn{4}{l}{\textit{\textbf{autoregressive/masking}}} \\
~~JetFormer-L~\cite{jetformer}     & 2.75B & 256$\times$2 & 6.64 \\
~~MaskGIT~\cite{maskgit}               & 227M & 8 & 6.18 \\
~~RQ-Transformer~\cite{RQT}   & 3.8B & 256$\times$2 & 3.80 \\
~~STARFlow~\cite{starflow}                & 1.4B & 1024$\times$2 & 2.40 \\
~~LLamaGen-3B~\cite{llamagen}      & 3.1B & 256$\times$2 & 2.18 \\
~~VAR-$d30$~\cite{var}               & 2B   & 10$\times$2 & 1.92 \\
~~MAR-H~\cite{mar}             & 943M & 256$\times$2 & 1.55 \\
~~RAR-XXL~\cite{rar}               & 1.5B & 256$\times$2 & 1.48\\
~~xAR-H~\cite{xar}                    & 1.1B & 50$\times$2 & 1.24 \\
\bottomrule
\end{tabular}%
}
}
\vspace{-.5em}
\caption{\textbf{System-level comparison on class-conditional ImageNet 256${\times}$256}.
\textbf{Left}: 1-NFE and 2-NFE diffusion/flow models trained \emph{from scratch}.
\textbf{Middle}: Diffusion/flow models, including distillation-based 1-NFE methods and multi-NFE methods.
\textbf{Right}: Reference methods from other generative modeling families, including GANs and autoregressive/masking models.
All numbers are with CFG when applicable, and $\times 2$ in NFE indicates that the CFG computation doubles NFEs at inference time.
}
\label{tab:all_results}
\vspace{-.25em}
\end{table*}

%% file: tables/visuals.tex
\begin{figure}[t]
\centering
\begin{minipage}[t]{\linewidth}
    \centering
    \includegraphics[width=\textwidth]{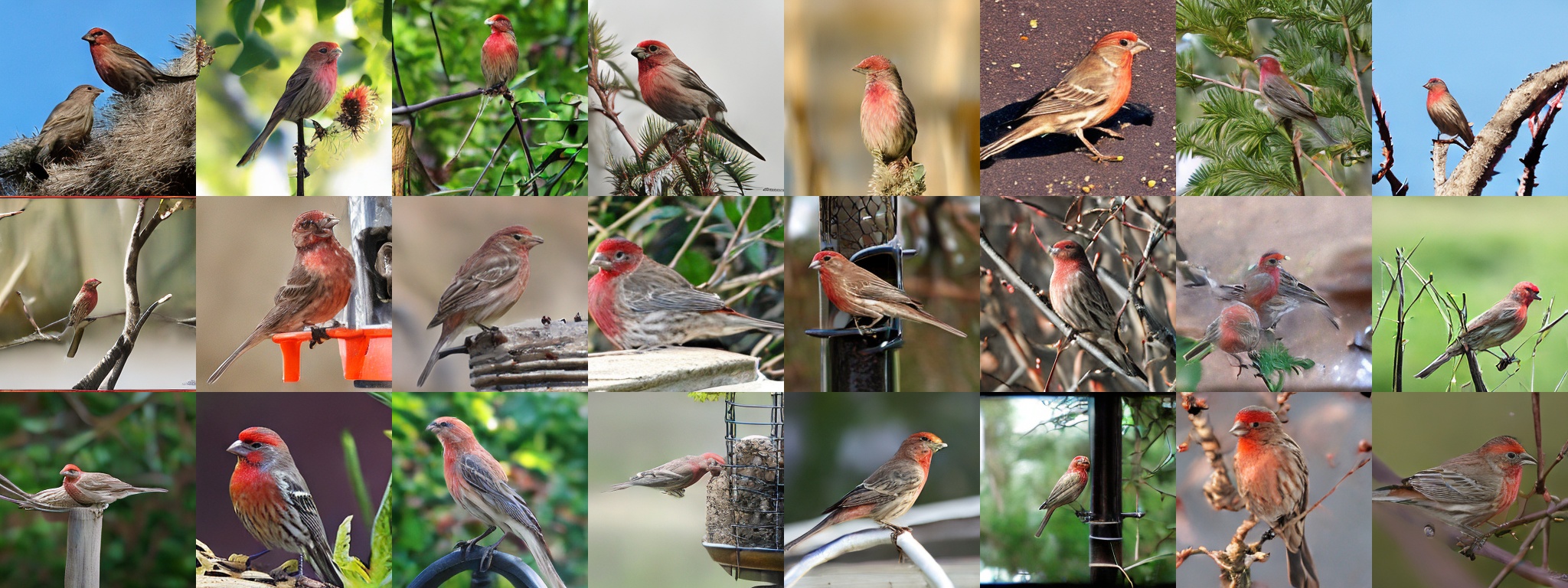}
    \\ \vspace{-0.25em}
    {\scriptsize class 12: house finch, linnet, Carpodacus mexicanus}
    \vspace{0.4em}
\end{minipage}

\begin{minipage}[t]{\linewidth}
    \centering
    \includegraphics[width=\textwidth]{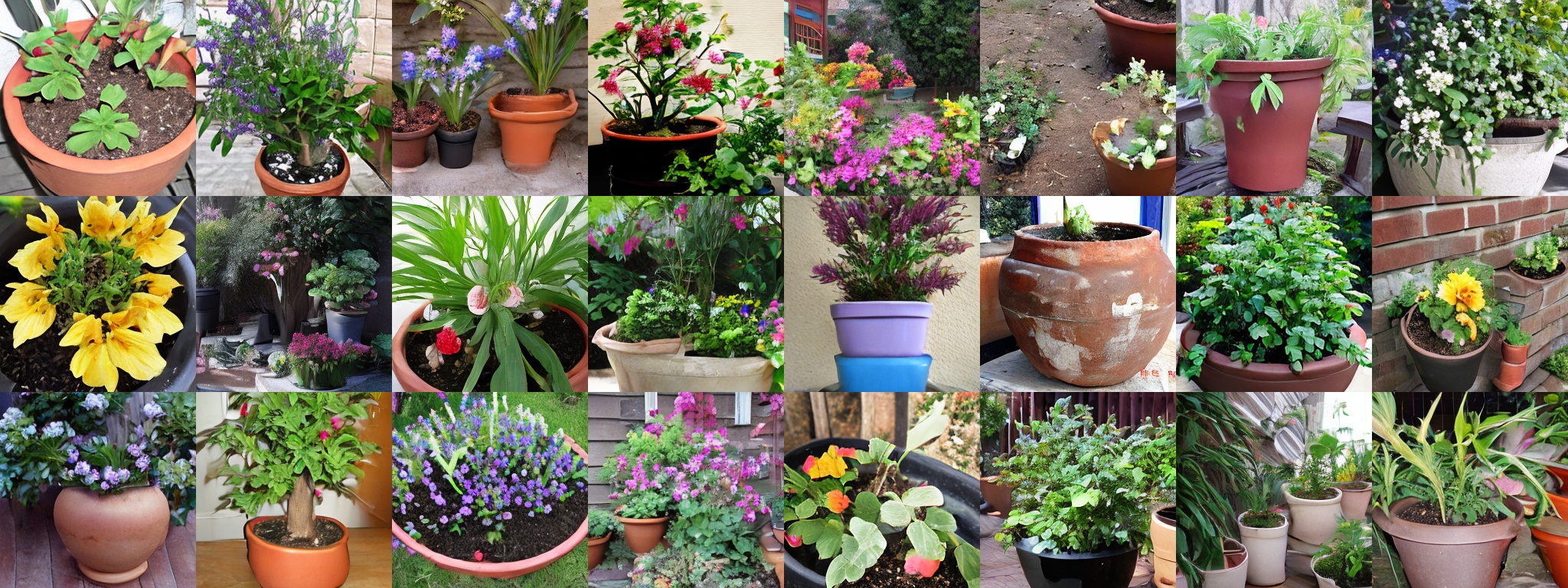}
    \\ \vspace{-0.25em}
    {\scriptsize class 738: pot, flowerpot}
    \vspace{0.4em}
\end{minipage}

\begin{minipage}[t]{\linewidth}
    \centering
    \includegraphics[width=\textwidth]{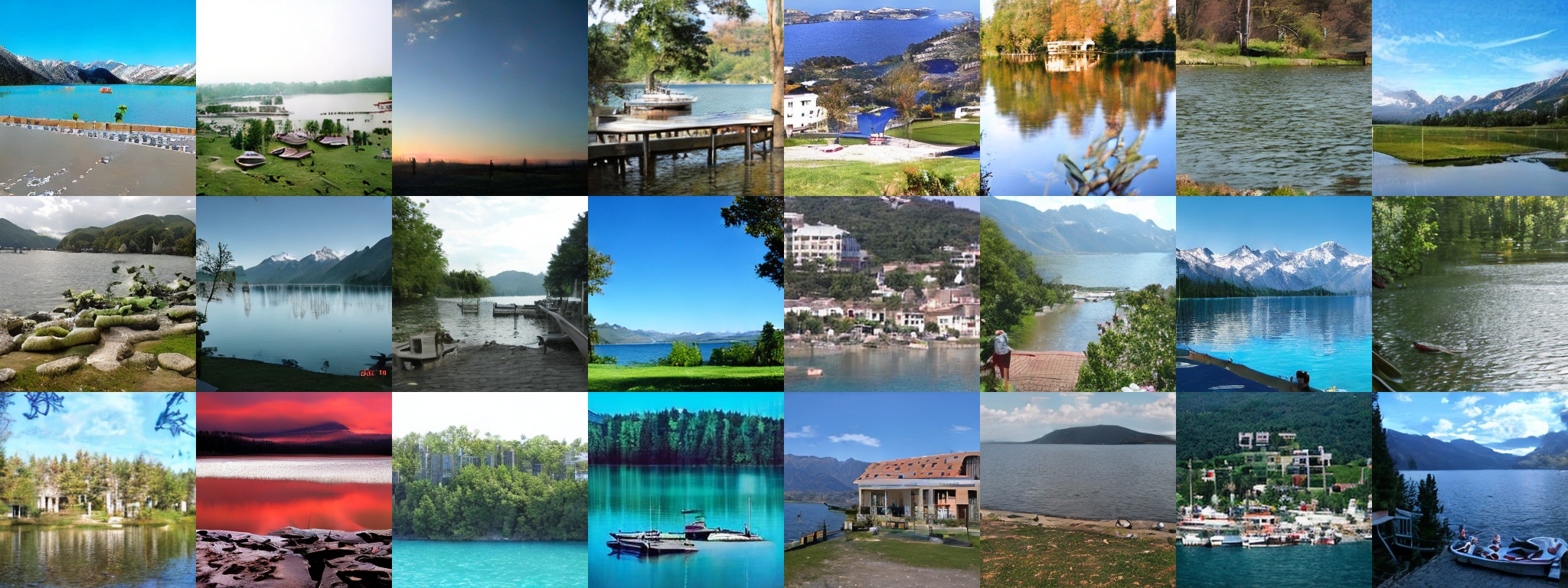}
    \\ \vspace{-0.25em}
    {\scriptsize class 975: lakeside, lakeshore}
\end{minipage}

\vspace{-0.5em}
\caption{\textbf{Qualitative results of 1-NFE generation on ImageNet 256$\times$256.}
We show \textit{uncurated} results on the three classes listed here; more are in appendix. The model is iMF-XL/2.
}
\vspace{-0.6em}
\label{fig:result_images}
\end{figure}

%% file: sections/conclusion.tex
\section{Conclusion}

We have demonstrated that fastforward generative models, without pretraining, can achieve highly competitive performance. We hope this encouraging result represents a solid step towards stand-alone fastforward generation.

With the remarkable progress of 1-NFE generation, the use of a tokenizer begins to incur a non-negligible cost at inference time. 
While our work focuses on advancing fast-forward models and is orthogonal to tokenizer design, from a practical standpoint, reducing or removing the tokenizer is becoming increasingly valuable.
We expect future research to explore efficient tokenizers or pixel-space generation.

%% file: sections/appendix.tex
\clearpage
\appendix

\section{Implementation Details}
\label{app:exp-settings}

\input{tables/settings}

The configurations and hyper-parameters are summarized in \cref{tab:imagenet-configs}. Our implementation is based on the public codebase of original MF, which is based on JAX and TPUs.\footnotemark{}

\footnotetext{\url{https://github.com/Gsunshine/meanflow}}

\paragraph{Auxiliary head for $v_\theta$.} In \cref{sec:v-pred}, we have introduced an auxiliary head for modeling $v_\theta$, which produces the input to the $\mathtt{JVP}$ computation. This auxiliary head shares most layers and computation with the main network $u_\theta$ and only differs in the last $L$ layers (we set $L=8$ in all cases). The output of this auxiliary head is to predict the marginal $v$.

Unlike the variant using the boundary condition of $u$ (that is, $v_\theta(z_t)\,=u_\theta(z_t, t, t)$), the unshared layers in the auxiliary head receive no gradient if not attached to any loss. To address this issue, we append an auxiliary loss $\|v_\theta - (e-x)\|^2$ to this head, which is the Flow Matching loss. The output of this head is only for the $\mathtt{JVP}$ computation and is not used at inference time.
As adaptive weighting~\cite{mf} is used in the MF loss, we also apply it to this auxiliary loss.

\paragraph{CFG conditioning.} In \cref{sec:method-cfg}, we have discussed CFG using the standard form \cite{cfg}: $v_{\text{cfg}} = \omega\,v(z_t\mid\mathbf{c}) + (1-\omega)\,v(z_t \mid \varnothing)$ (\cref{eq:cfg}).
The original MF paper \cite{mf} derives a relationship between $v_{\text{cfg}}$ and its resulting average velocity field, which can be simplified as (see Eq.~(21) in \cite{mf}):
\begin{equation}
    v_{\text{tgt}}=(e-x)+(1-\frac{1}{\omega})\big(u^\text{cfg}_\theta(z_t\mid t,t,\mathbf{c}) - u^\text{cfg}_\theta(z_t\mid t,t, \varnothing)\big),
\end{equation}
where ``$\varnothing$'' is to emphasize the unconditional field. 
Here, $\omega$ is the ``effective guidance scale'' \cite{mf}, which plays the same role as in standard CFG.
Specifically, when $\omega=1$, this equation degenerates to the no-CFG case. 
We adopt this formulation.
See the pseudocode in \cref{alg:cfg_code}. 

\input{sections/cfg_code}

When using CFG-conditioning, we need to randomly sample the scale $\omega$ for each training sample.
First, we set a sampling range of $\omega$:  $[1.0,\, \omega_{\max}]$, where we fix $\omega_{\max}=8.0$.
Note that $\omega=1$ degenerates to the no-CFG case.
Then we sample $\omega$ from a power distribution that biases towards smaller $\omega$ values: $\omega \sim p(\omega)\propto\omega^{-\beta}$,
where $\beta$ controls the skewness (we use $\beta = 1$ or $2$, \cref{tab:imagenet-configs}).

When using CFG-conditioning to support guidance interval \cite{interval}, we randomly sample $t_{\min}$ and $t_{\max}$ from $\mathcal{U}[0,0.5]$ and $\mathcal{U}[0.5,1.0]$ respectively, where $\mathcal{U}$ is the uniform distribution. During training, when $t$ falls outside of $[t_{\min}, t_{\max}]$, CFG is turned off by setting $\omega=1$.
The set of sampled values of $\Omega=\{\omega, t_{\min}, t_{\max}\}$ is provided to the network as extra conditioning.

\paragraph{In-context conditioning.} Our models are conditioned on time steps $r, t$, class $\mathbf{c}$, and CFG factors $\Omega=\{\omega, t_{\min}, t_{\max}\}$. All continuous-valued conditions (\eg $\omega$) are processed by standard positional embedding \cite{transformer}, similar to how $t$-conditioning is handled in continuous-time diffusion/flow models.
Each type of these conditions is processed by a 2-layer MLP. All conditions are replicated into multiple tokens: the number of replications for each type is in \cref{tab:imagenet-configs}. All replicated tokens are added with learnable embeddings indicating their types of conditions (analogous to position embedding along the sequence), and then are concatenated along the sequence axis with the image tokens (see \cref{fig:network}).

\paragraph{Removing adaLN-zero.} When using in-context conditioning, our model removes the standard adaLN-zero \cite{dit} that is parameter-heavy.
We adopt the zero residual-block initialization \cite{largesgd}, which adaLN-zero \cite{dit} also follows. Specifically, for any residual block \cite{resnet} in a Transformer \cite{transformer} with the form of $x+F(x)$, the last operation in $F(x)$ is always a learnable per-channel scale $\gamma$, where $\gamma$ is initialized as zero. As such, the initial state of a residual block is always identity mapping. The initialization of adaLN-zero \cite{dit} was based on the same principle.

For all other linear projection layers in the Transformer blocks, we use a Gaussian initialization $\mathcal{N}(0, \sigma^2)$, where $\sigma^2=0.1 / \mathtt{fan\_in}$ and $\mathtt{fan\_in}$ is the input channel number. This can be implemented in popular libraries by $\sigma=\mathtt{gain} / \sqrt{\mathtt{fan\_in}}$ where $\mathtt{gain}=\sqrt{0.1}\approx 0.32$. This initialized $\sigma$ is more conservative than common gain-controlled initializations, where $\mathtt{gain}$ is 1 or $\sqrt{2}$. In our preliminary experiments, this initialization converges faster when our block becomes different from the adaLN-zero block.

\paragraph{Evaluation}.
We sample 50,000 samples and compute FID against the ImageNet training set (\ie FID-50K). We sample 50 images per class for the FID evaluation. For each of our models where CFG-conditioning is enabled, we report the FID results using the optimal guidance scale and interval.

\section{Additional Qualitative Results} 
\label{app:vis}
We provide additional qualitative results in \cref{fig:appendix_gen1} and \cref{fig:appendix_gen2}.
These results are uncurated samples of the classes listed as conditions.
These results (and \cref{fig:result_images}) are from our iMF-XL/2 model for 1-NFE ImageNet 256$\times$256 generation.
Following common practice, we present qualitative results using a CFG setting that favors the IS metric (emphasizing individual quality) at the expense of the FID metric (emphasizing diversity and distributional coverage); note that this tradeoff was impossible in the original MF, where the CFG is fixed.
Here, we set CFG as $\omega=6.0$ and CFG interval as $[t_\text{min}, t_\text{max}]=[0.2,0.8]$. This evaluation setting has an FID of 3.92 and an IS of 348.2.

%% file: tables/settings.tex
\begin{table}[t]
\small
\centering
\begin{adjustbox}{max width=\linewidth}
\begin{tabular}{y{60}x{40}x{28}x{28}x{33}}
\toprule
configs & iMF-B & iMF-M & iMF-L & iMF-XL \\
\midrule
params (M)
            & 89  & 174 & 409  & 610 \\
depth
            & 12  & 24  & 32   & 48 \\
hidden dim
            & 768 & 768 & 1024 & 1024 \\
attn heads
            & 12  & 12  & 16   & 16 \\
patch size
            & \multicolumn{4}{c}{$2{\times}2$} \\
aux-head depth  
            & \multicolumn{4}{c}{8} \\
class tokens 
            & \multicolumn{4}{c}{8} \\
time tokens 
            & \multicolumn{4}{c}{4} \\
guidance tokens  
            & \multicolumn{4}{c}{4} \\
interval tokens  
            & \multicolumn{4}{c}{4} \\
linear layer init
            & \multicolumn{4}{c}{ $\mathcal{N}(0, \sigma^2)$, $\sigma^2=0.1 / \mathtt{fan\_in}$ } \\
\midrule
epochs
            & 240$^\dagger$ / 640 & 640 & 640 & 800    \\
batch size  
            & \multicolumn{4}{c}{256$^\dagger$ / 1024} \\
learning rate 
            & \multicolumn{4}{c}{0.0001} \\
lr schedule
            & \multicolumn{4}{c}{constant} \\
lr warmup~\cite{largesgd} 
            & \multicolumn{4}{c}{10 epochs} \\
optimizer
            & \multicolumn{4}{c}{Adam \cite{adam}} \\
Adam $(\beta_1, \beta_2)$
             & \multicolumn{4}{c}{(0.9, 0.95)} \\
weight decay 
            & \multicolumn{4}{c}{0.0} \\
dropout
            & \multicolumn{4}{c}{0.0} \\
ema decay 
            & \multicolumn{4}{c}{0.9999} \\
\midrule
ratio of $r{\neq}t$ 
            & \multicolumn{4}{c}{50\%} \\
$(t,r)$ cond 
            & \multicolumn{4}{c}{$t-r$} \\
$t,r$ sampler 
            & \multicolumn{4}{c}{logit-normal($-0.4$, $1.0$)} \\
\midrule
cls drop \cite{cfg} 
            & \multicolumn{4}{c}{0.1} \\
CFG dist $\beta$ 
            & 1 & 2 & 2 & 2 \\ 
\bottomrule
\end{tabular}
\end{adjustbox}
\caption{\textbf{Configurations and hyper-parameters}. $^\dagger$: these are for ablation studies.}
\label{tab:imagenet-configs}
\end{table}

%% file: sections/cfg_code.tex
\definecolor{codeblue}{rgb}{0.25,0.5,0.5}
\definecolor{codesign}{RGB}{0, 0, 255}
\definecolor{codefunc}{rgb}{0.85, 0.18, 0.50}

\lstdefinelanguage{PythonFuncColor}{
  language=Python,
  keywordstyle=\color{blue}\bfseries,
  commentstyle=\color{codeblue},
  stringstyle=\color{orange},
  showstringspaces=false,
  basicstyle=\ttfamily\small,
  literate=
    {+}{{\color{codesign}+ }}{1}
    {-}{{\color{codesign}- }}{1}
    {*}{{\color{codesign}* }}{1}
    {/}{{\color{codesign}/ }}{1}
    {sample_t_r_cfg}{{\color{codefunc}sample\_t\_r\_cfg}}{1}
    {randn_like}{{\color{codefunc}randn\_like}}{1}
    {jvp}{{\color{codefunc}jvp}}{1}
    {stopgrad}{{\color{codefunc}stopgrad}}{1}
    {metric}{{\color{codefunc}metric}}{1}
}

\lstset{
  language=PythonFuncColor,
  backgroundcolor=\color{white},
  basicstyle=\fontsize{8pt}{8.8pt}\ttfamily\selectfont,
  columns=fullflexible,
  breaklines=true,
  captionpos=b,
}

\begin{algorithm}[t]
\centering
\caption{{improved MeanFlow}: training guidance.\\
{\scriptsize Note: in PyTorch and JAX, \texttt{jvp} returns the function output and JVP.}}

\begin{minipage}{0.99\linewidth}
\begin{lstlisting}[language=PythonFuncColor, escapechar=`]
# fn(z, t, r, w, c): function to predict u
# x: training batch
# c: condition batch

t, r, w = sample_t_r_cfg()
e = randn_like(x)

z = (1 - t) * x + t * e

# cls cond and cls uncond v
v_c = fn(z, t, t, w, c)
v_u = fn(z, t, t, w, None)

# Compute CFG target (same as orig MF)
v_tgt = (e - x) + (1 - 1 / w) * (v_c - v_u)

# Use predicted v_c to compute dudt
u, dudt = jvp(fn,  (z, r, t, w, c),  
                        (v_c, 0, 1, 0, 0))

# Compute compound function V
V = u + (t - r) * stopgrad(dudt)
error = V - stopgrad(v_tgt)

loss = metric(error)
\end{lstlisting}
\end{minipage}
\label{alg:cfg_code}
\end{algorithm}

%% file: sections/acknowledgement.tex
\vspace{2em}
\paragraph{Acknowledgment.}
We greatly thank Google TPU Research Cloud (TRC) for granting us access to TPUs.
Zhengyang Geng is partially supported by funding from the Bosch Center for AI. Zico Kolter gratefully acknowledges Bosch’s funding for the lab.
We thank Hanhong Zhao, Qiao Sun, Zhicheng Jiang and Xianbang Wang for their help on the JAX and TPU implementation. We thank our group members for helpful discussions and feedback.

%% file: sections/vis.tex
\begin{figure*}[t]
    \centering
    \begin{minipage}[t]{0.46\linewidth}
        \centering
        \includegraphics[width=\textwidth]{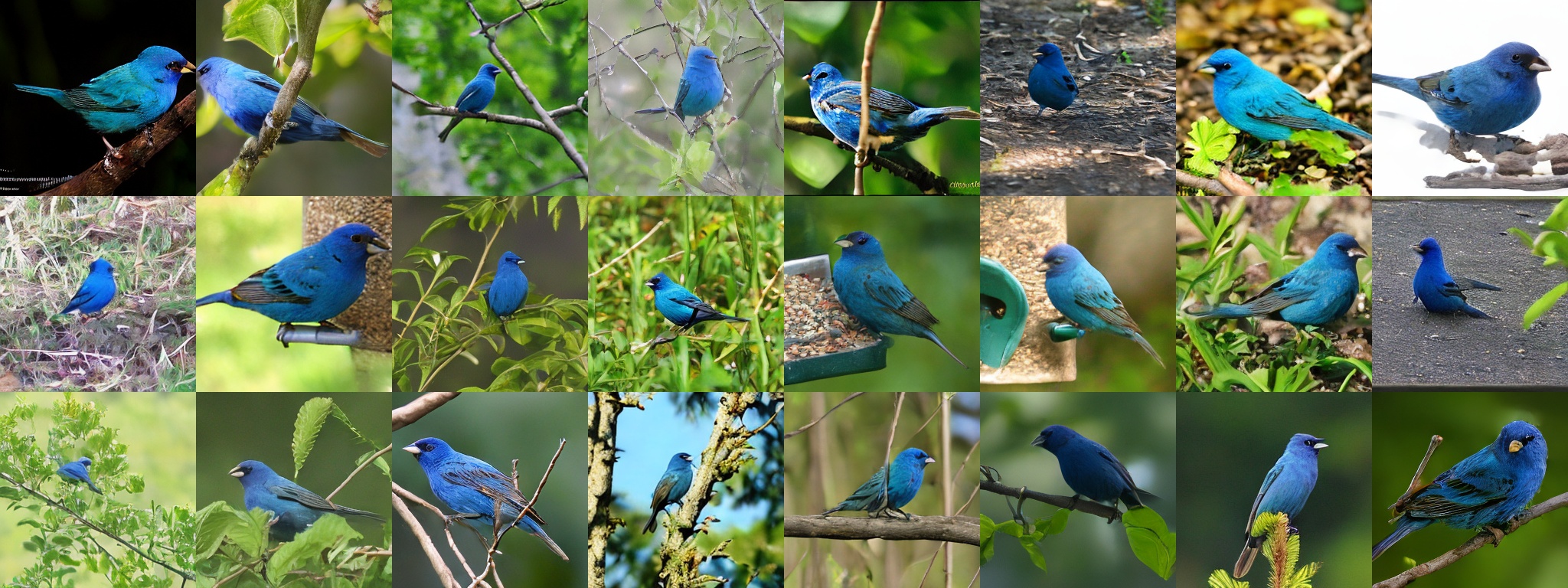}
        {\scriptsize class 14: indigo bunting, indigo finch, indigo bird, Passerina cyanea}
        \vspace{.5em}
    \end{minipage}
    \hfill
    \begin{minipage}[t]{0.46\linewidth}
        \centering
        \includegraphics[width=\textwidth]{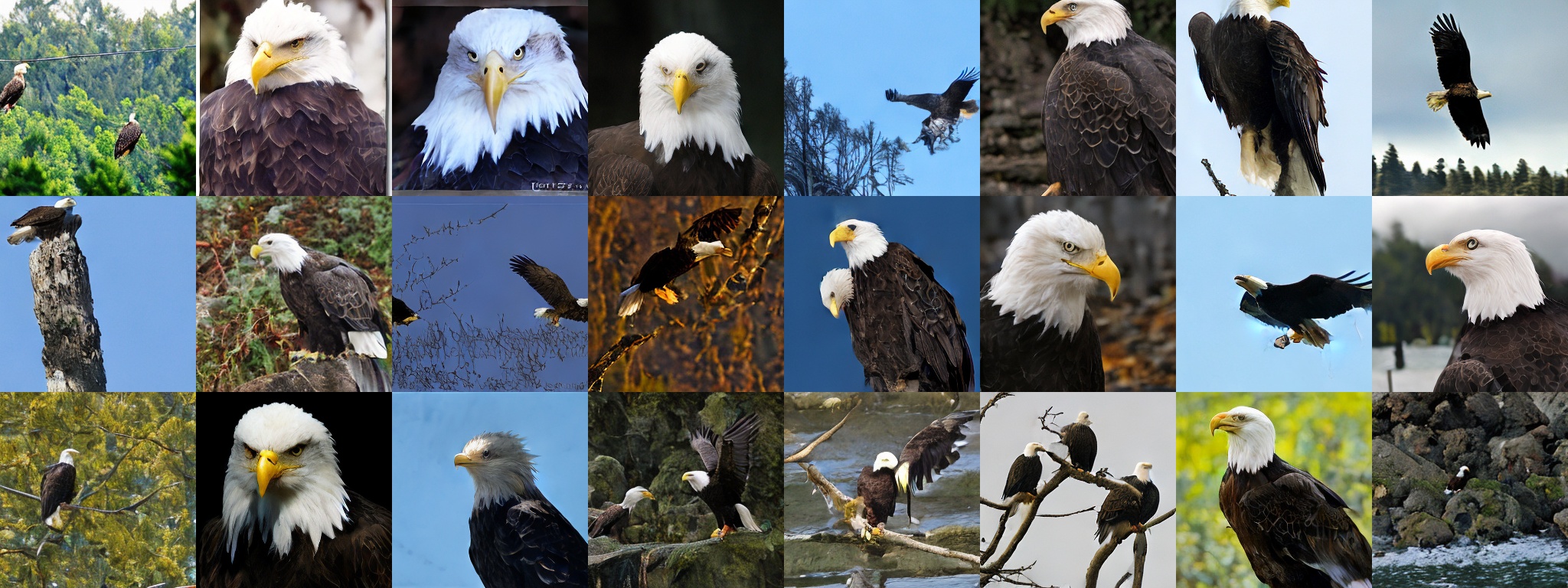}
        {\scriptsize class 22: bald eagle, American eagle, Haliaeetus leucocephalus}
        \vspace{.5em}
    \end{minipage}
    
    \begin{minipage}[t]{0.46\linewidth}
        \centering
        \includegraphics[width=\textwidth]{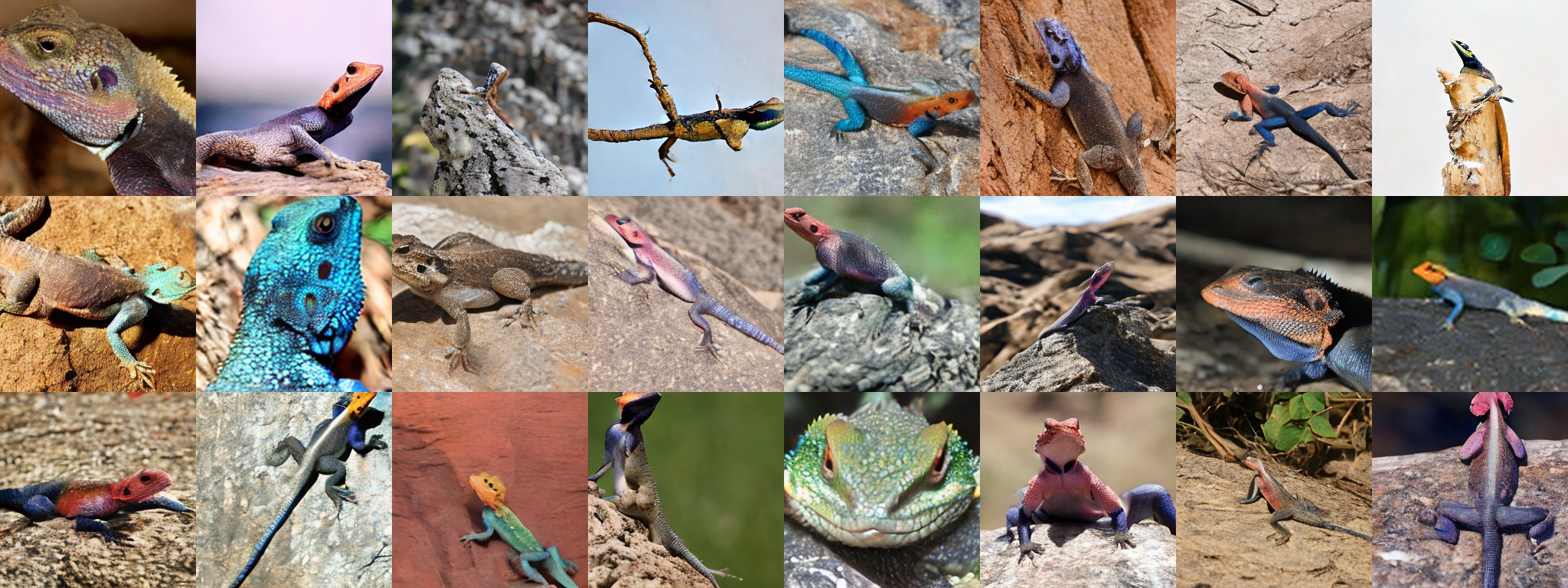}
        {\scriptsize class 42: agama}
        \vspace{.5em}
    \end{minipage}
    \hfill
    \begin{minipage}[t]{0.46\linewidth}
        \centering
        \includegraphics[width=\textwidth]{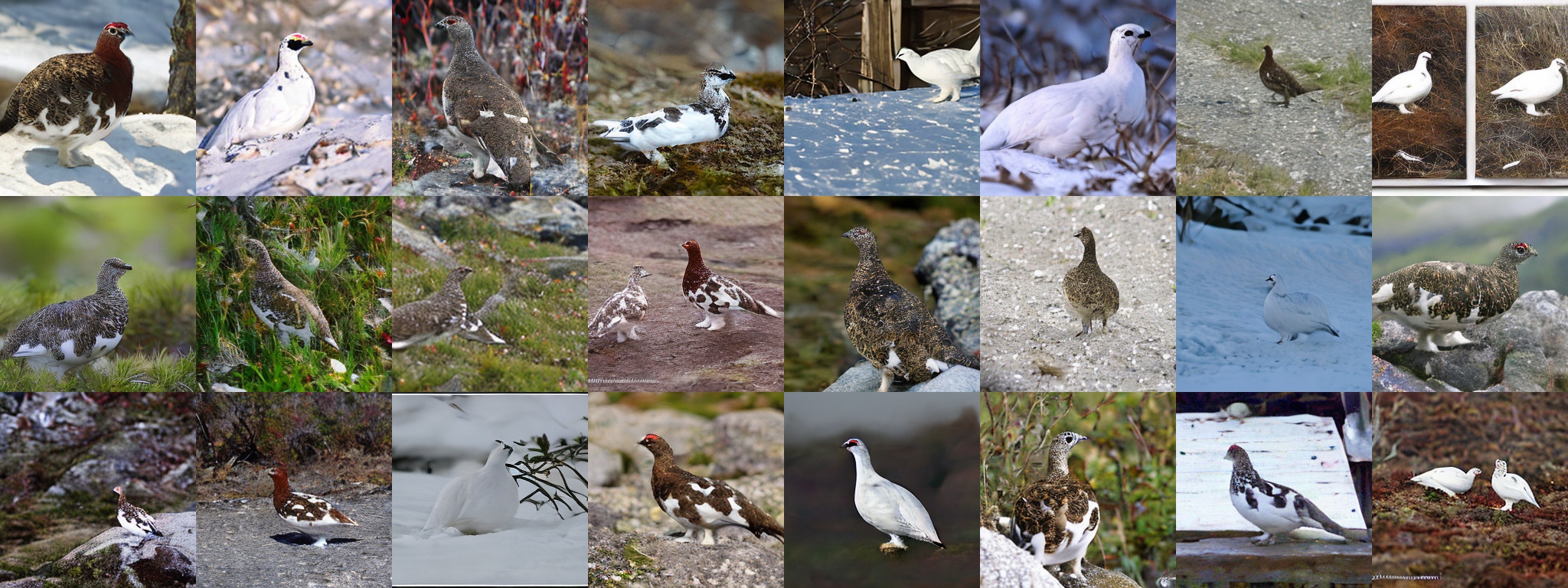}
        {\scriptsize class 81: ptarmigan}
        \vspace{.5em}
    \end{minipage}

    \begin{minipage}[t]{0.46\linewidth}
        \centering
        \includegraphics[width=\textwidth]{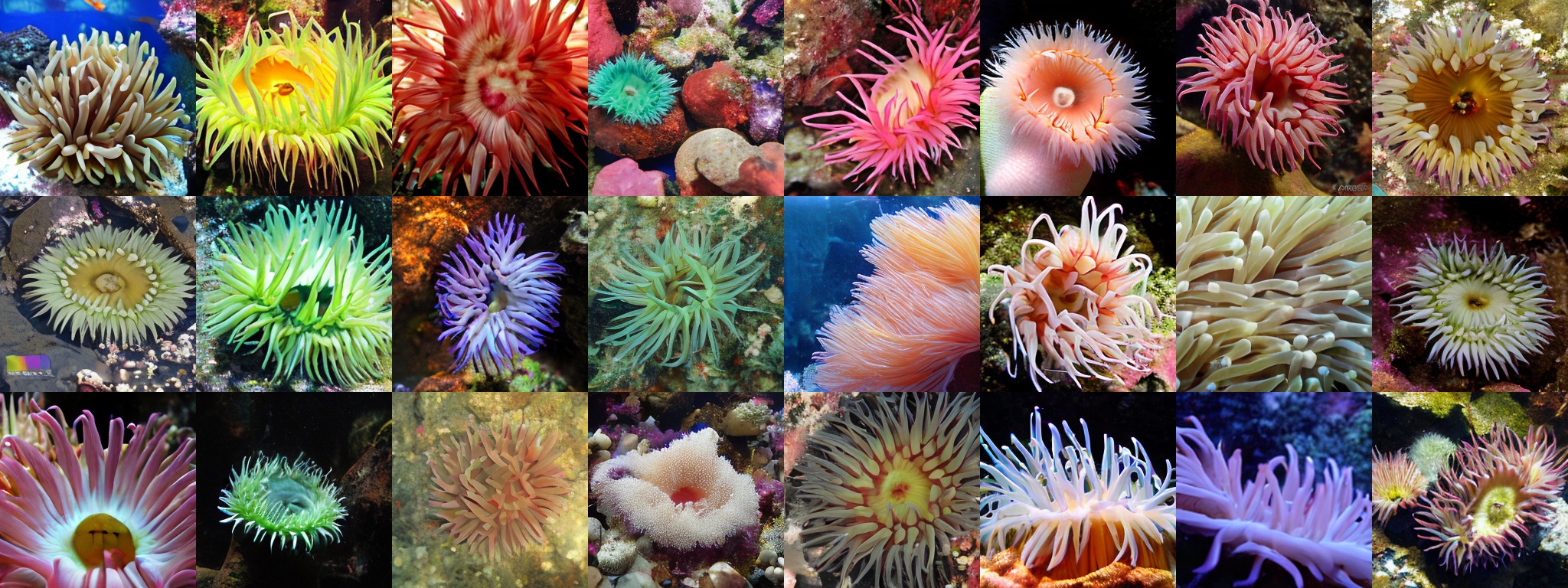}
        {\scriptsize class 108: sea anemone, anemone}
        \vspace{.5em}
    \end{minipage}
    \hfill
    \begin{minipage}[t]{0.46\linewidth}
        \centering
        \includegraphics[width=\textwidth]{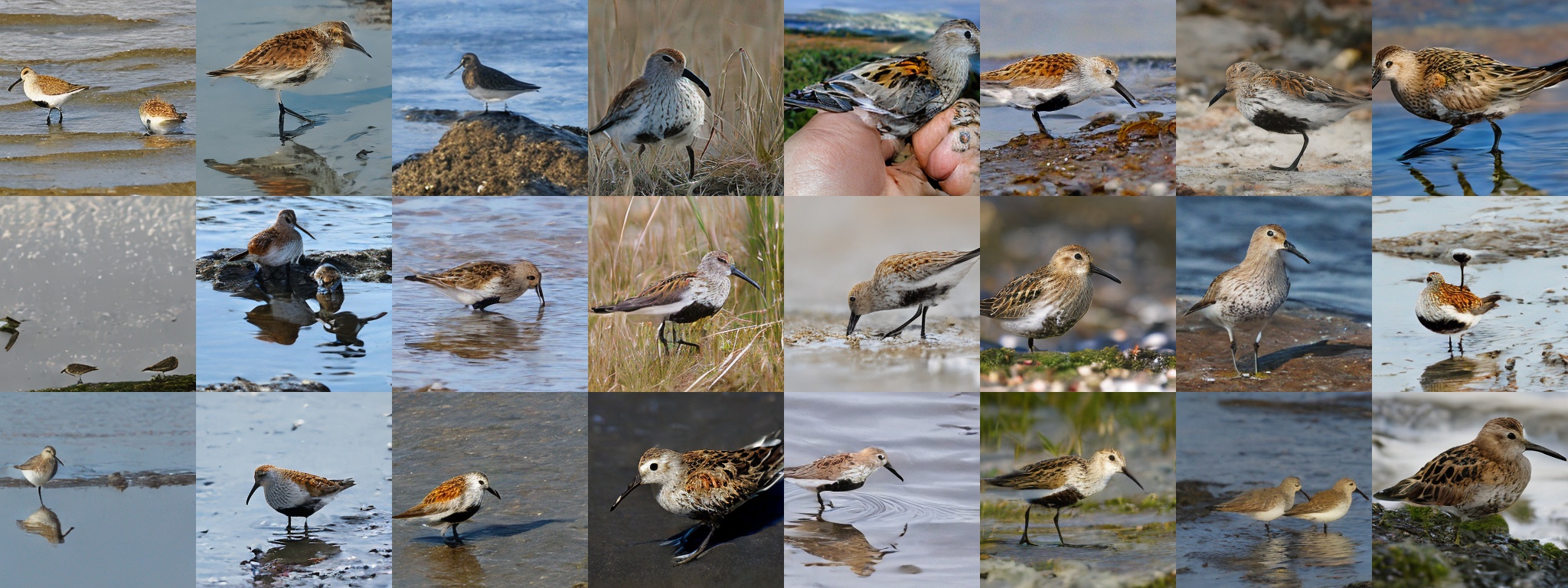}
        {\scriptsize class 140: red-backed sandpiper, dunlin, Erolia alpina}
        \vspace{.5em}
    \end{minipage}

    \begin{minipage}[t]{0.46\linewidth}
        \centering
        \includegraphics[width=\textwidth]{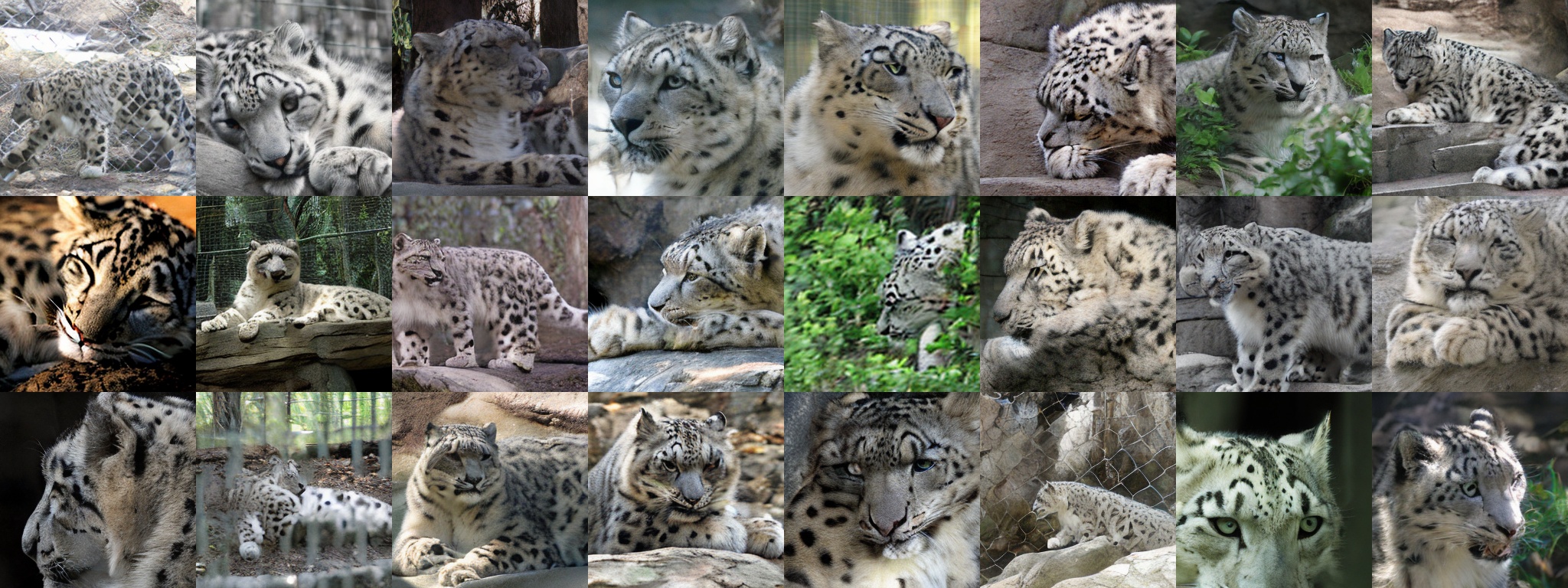}
        {\scriptsize class 289: snow leopard, ounce, Panthera uncia}
        \vspace{.5em}
    \end{minipage}
    \hfill
    \begin{minipage}[t]{0.46\linewidth}
        \centering
        \includegraphics[width=\textwidth]{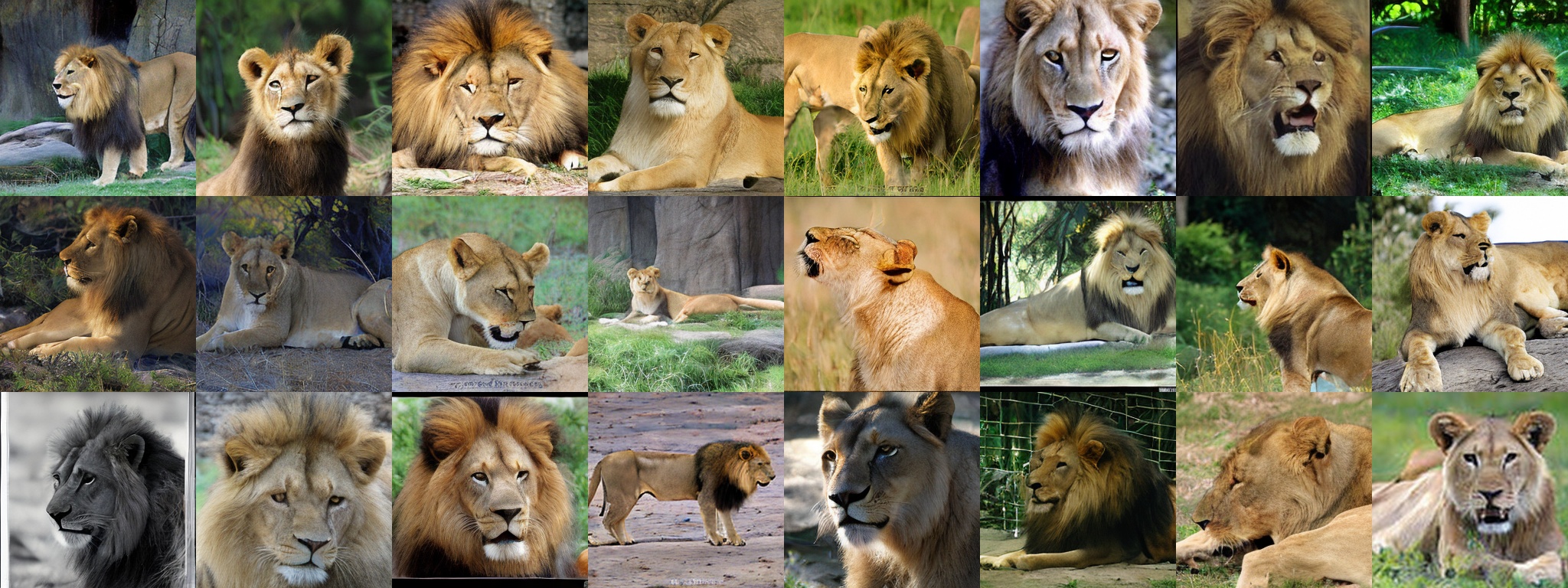}
        {\scriptsize class 291: lion, king of beasts, Panthera leo}
        \vspace{.5em}
    \end{minipage}

    \begin{minipage}[t]{0.46\linewidth}
        \centering
        \includegraphics[width=\textwidth]{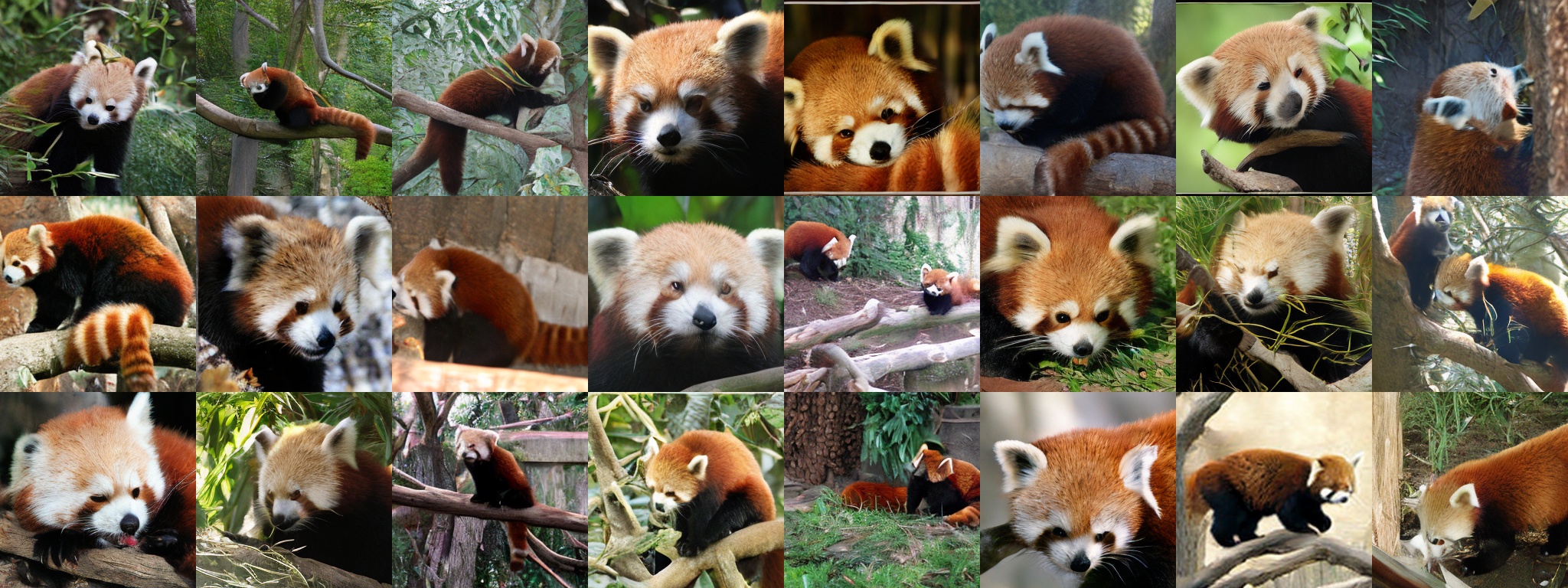}
        {\scriptsize class 387: lesser panda, red panda, panda, bear cat, cat bear, Ailurus fulgens}
        \vspace{.5em}
    \end{minipage}
    \hfill
    \begin{minipage}[t]{0.46\linewidth}
        \centering
        \includegraphics[width=\textwidth]{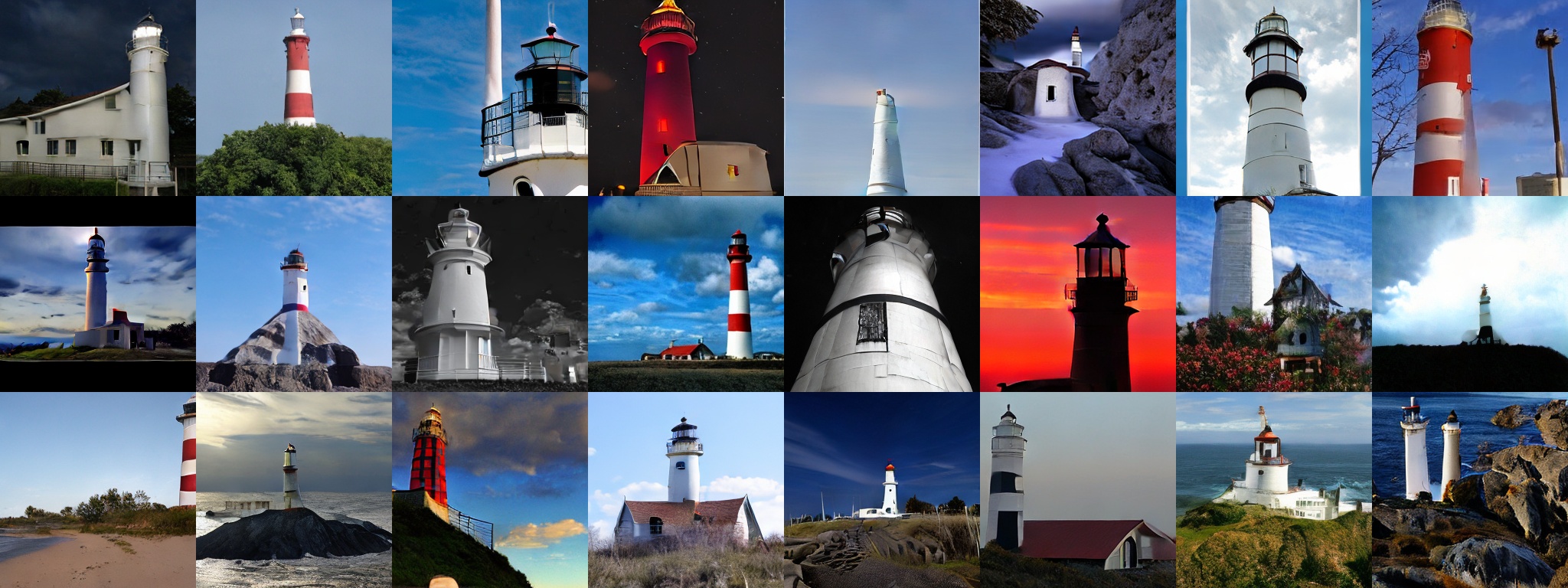}
        {\scriptsize class 437: beacon, lighthouse, beacon light, pharos}
        \vspace{.5em}
    \end{minipage}
    
    \vspace{-1em}
    \caption{\textit{Uncurated} 1-NFE class-conditional generation samples of iMF-XL/2 on ImageNet 256$\times$256.}
    \label{fig:appendix_gen1}
    \vspace{-1em}
\end{figure*}

\begin{figure*}[t]
    \centering

    \begin{minipage}[t]{0.46\linewidth}
        \centering
        \includegraphics[width=\textwidth]{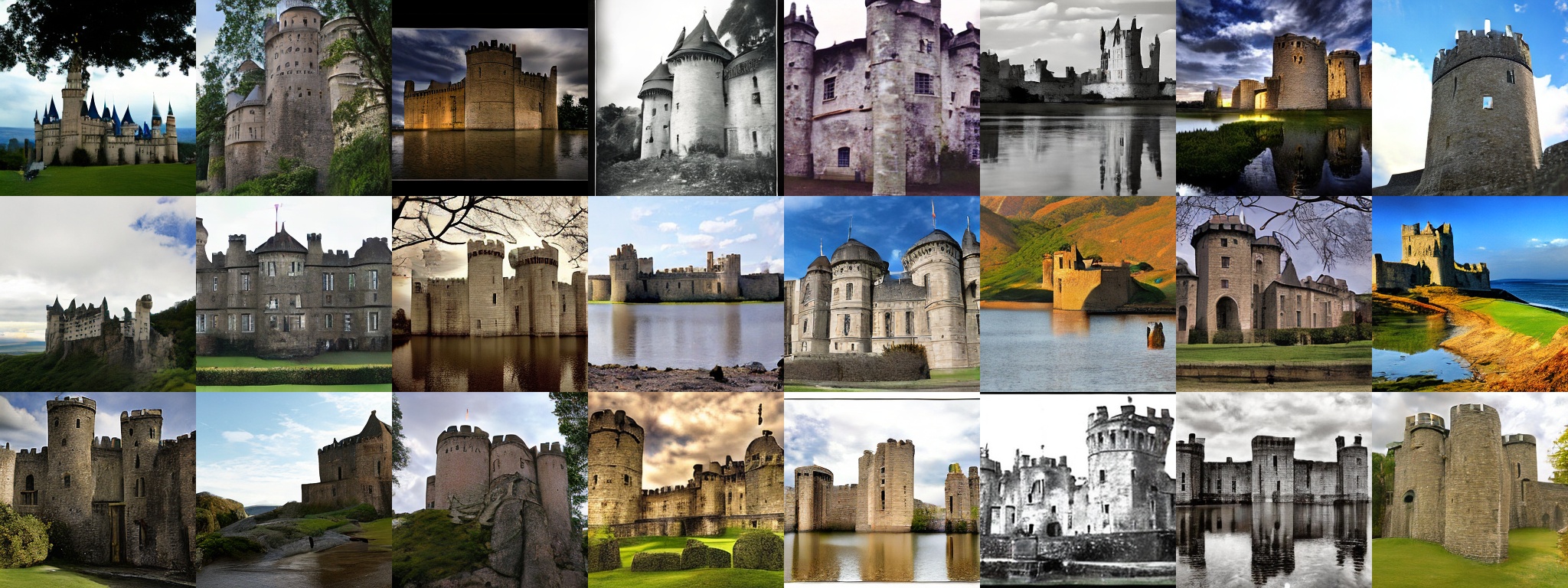}
        {\scriptsize class 483: castle}
        \vspace{.5em}
    \end{minipage}
    \hfill
    \begin{minipage}[t]{0.46\linewidth}
        \centering
        \includegraphics[width=\textwidth]{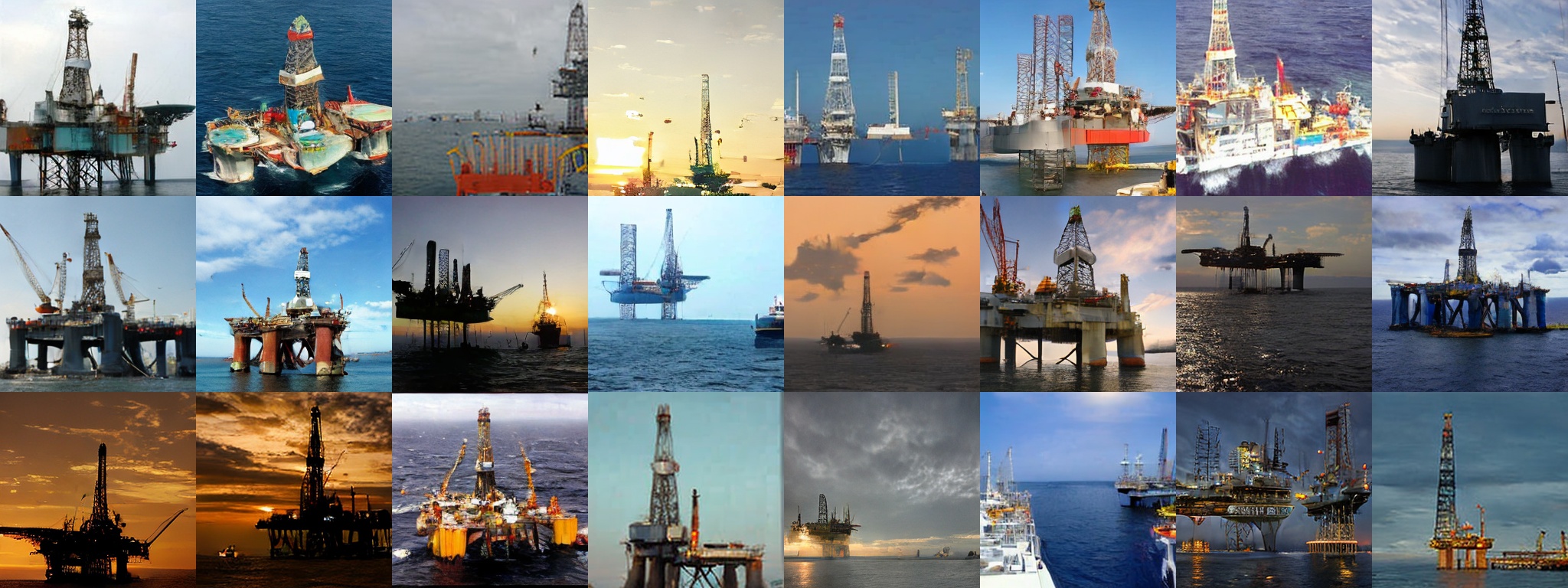}
        {\scriptsize class 540: drilling platform, offshore rig}
        \vspace{.5em}
    \end{minipage}

    \begin{minipage}[t]{0.46\linewidth}
        \centering
        \includegraphics[width=\textwidth]{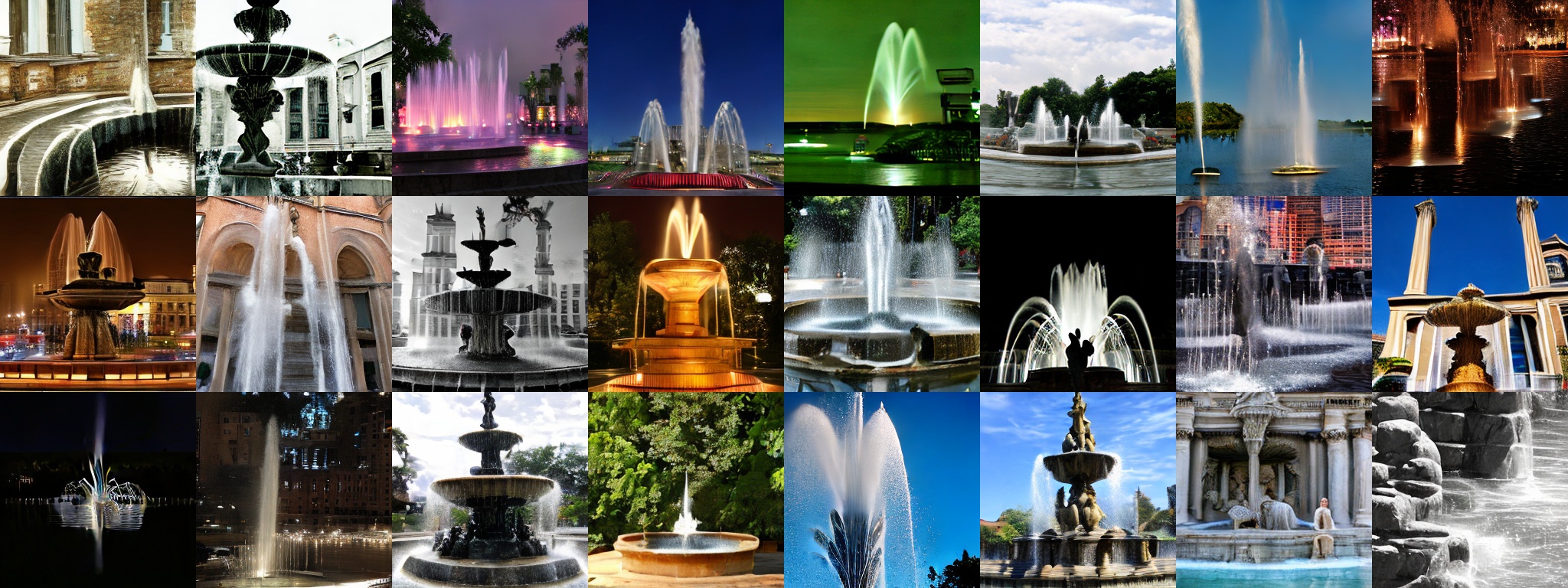}
        {\scriptsize class 562: fountain}
        \vspace{.5em}
    \end{minipage}
    \hfill
    \begin{minipage}[t]{0.46\linewidth}
        \centering
        \includegraphics[width=\textwidth]{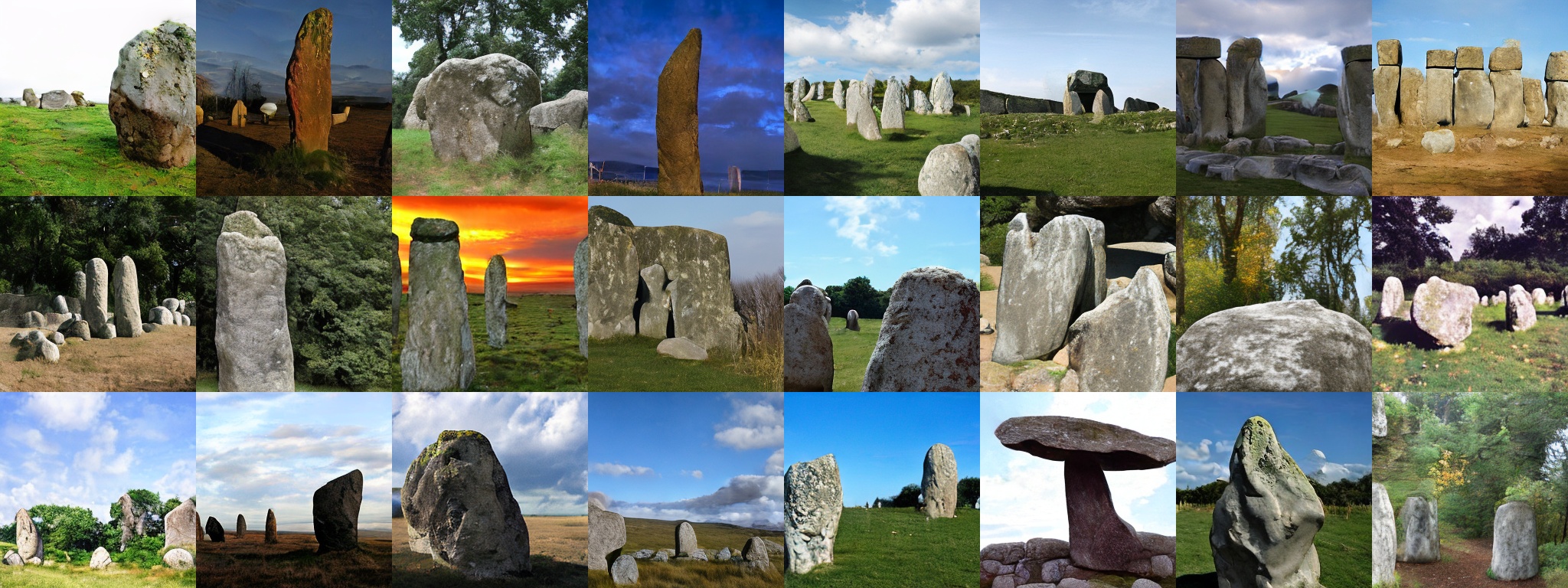}
        {\scriptsize class 649: megalith, megalithic structure}
        \vspace{.5em}
    \end{minipage}

    \begin{minipage}[t]{0.46\linewidth}
        \centering
        \includegraphics[width=\textwidth]{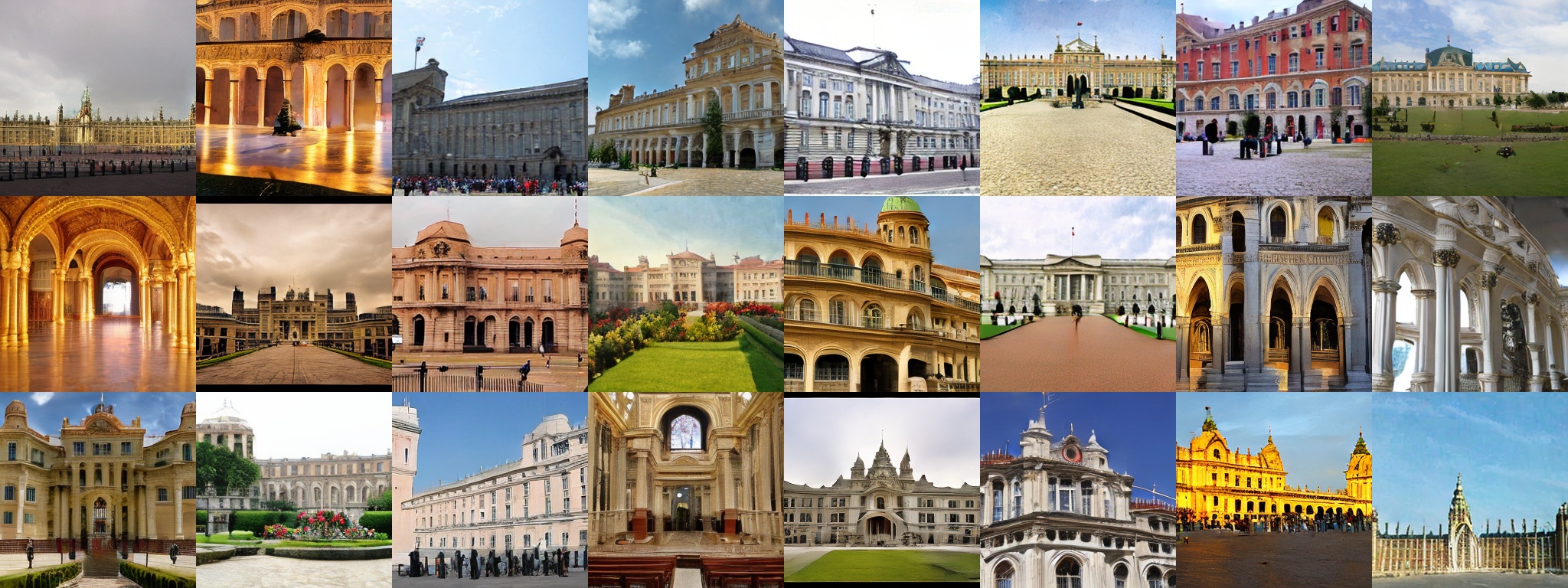}
        {\scriptsize class 698: palace}
        \vspace{.5em}
    \end{minipage}
    \hfill
    \begin{minipage}[t]{0.46\linewidth}
        \centering
        \includegraphics[width=\textwidth]{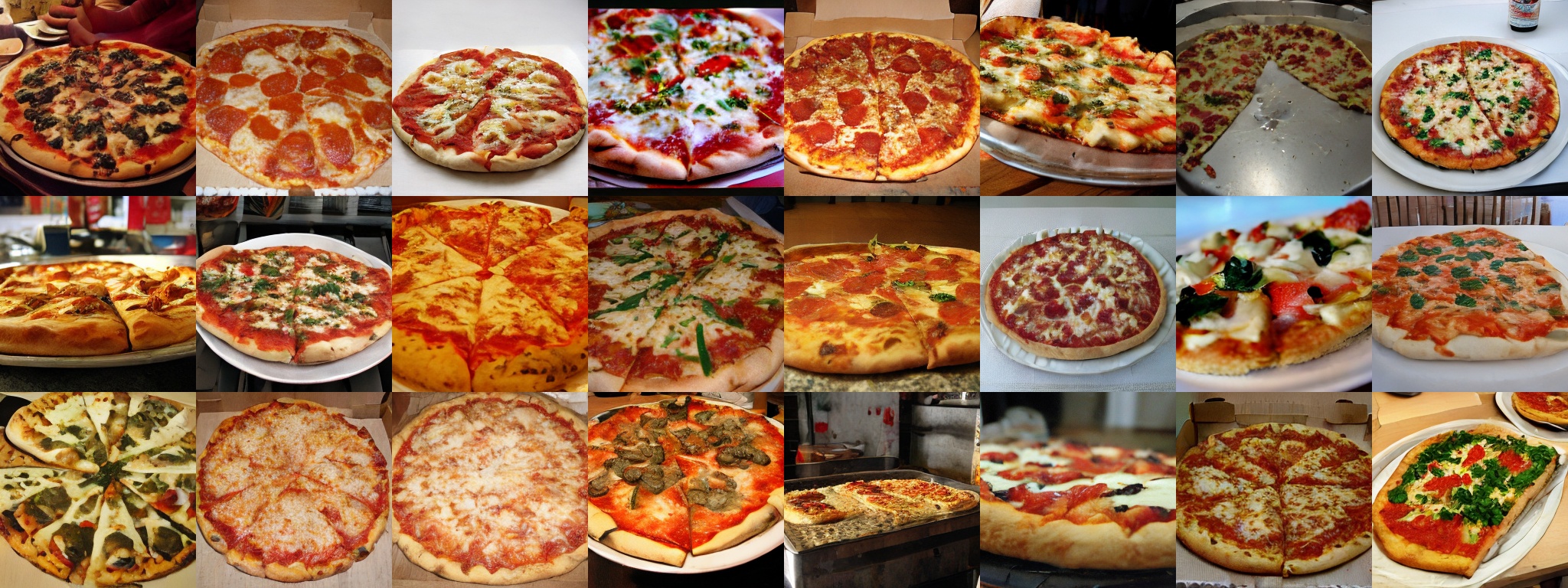}
        {\scriptsize class 963: pizza, pizza pie}
        \vspace{.5em}
    \end{minipage}

    \begin{minipage}[t]{0.46\linewidth}
        \centering
        \includegraphics[width=\textwidth]{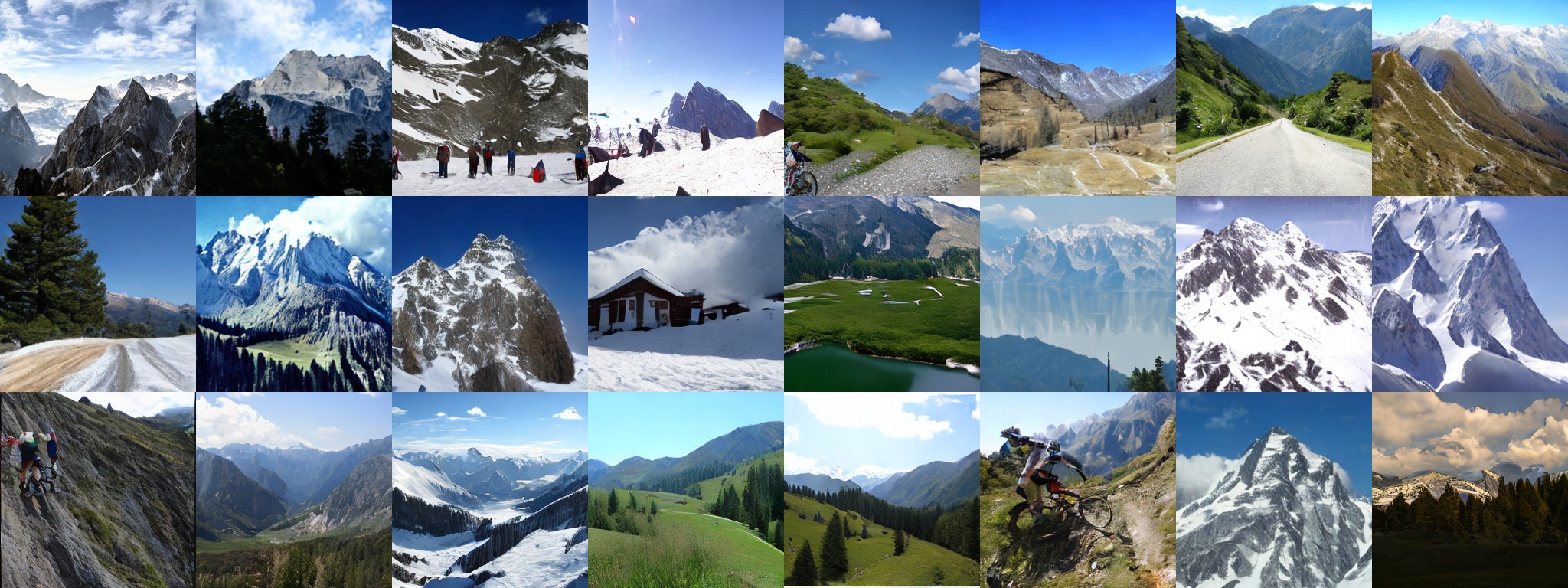}
        {\scriptsize class 970: alp}
        \vspace{.5em}
    \end{minipage}
    \hfill
    \begin{minipage}[t]{0.46\linewidth}
        \centering
        \includegraphics[width=\textwidth]{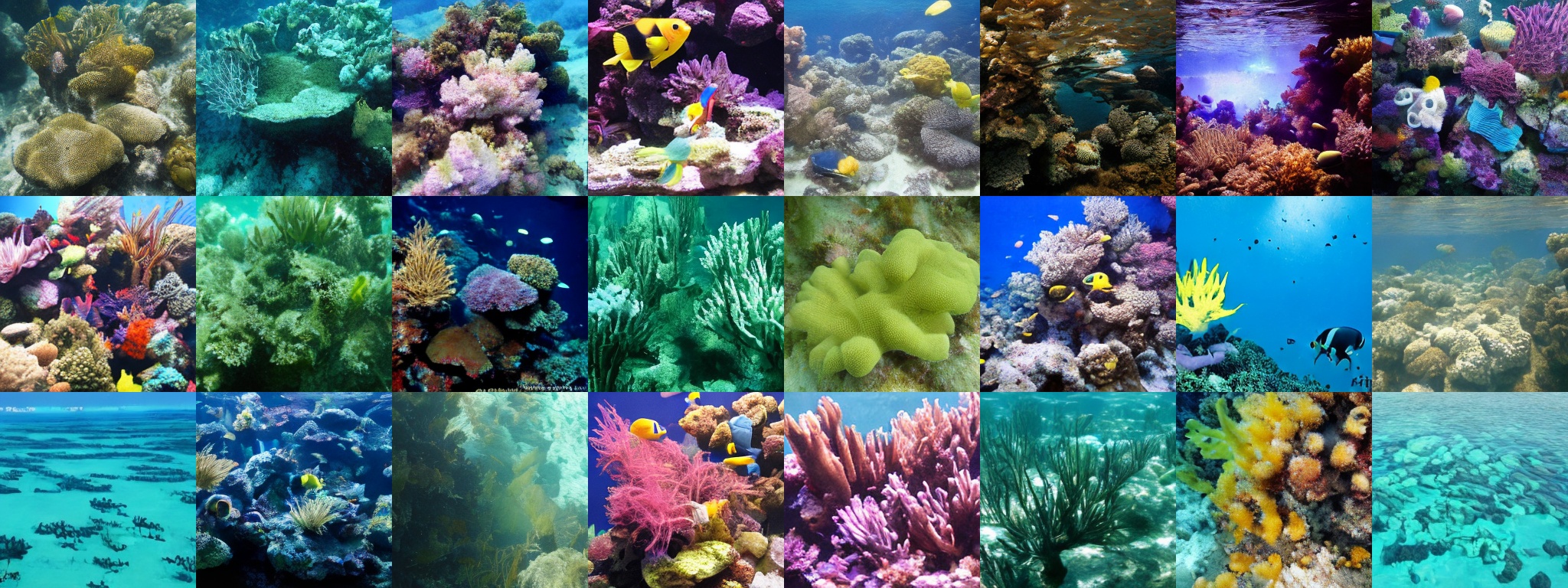}
        {\scriptsize class 973: coral reef}
        \vspace{.5em}
    \end{minipage}

    \begin{minipage}[t]{0.46\linewidth}
        \centering
        \includegraphics[width=\textwidth]{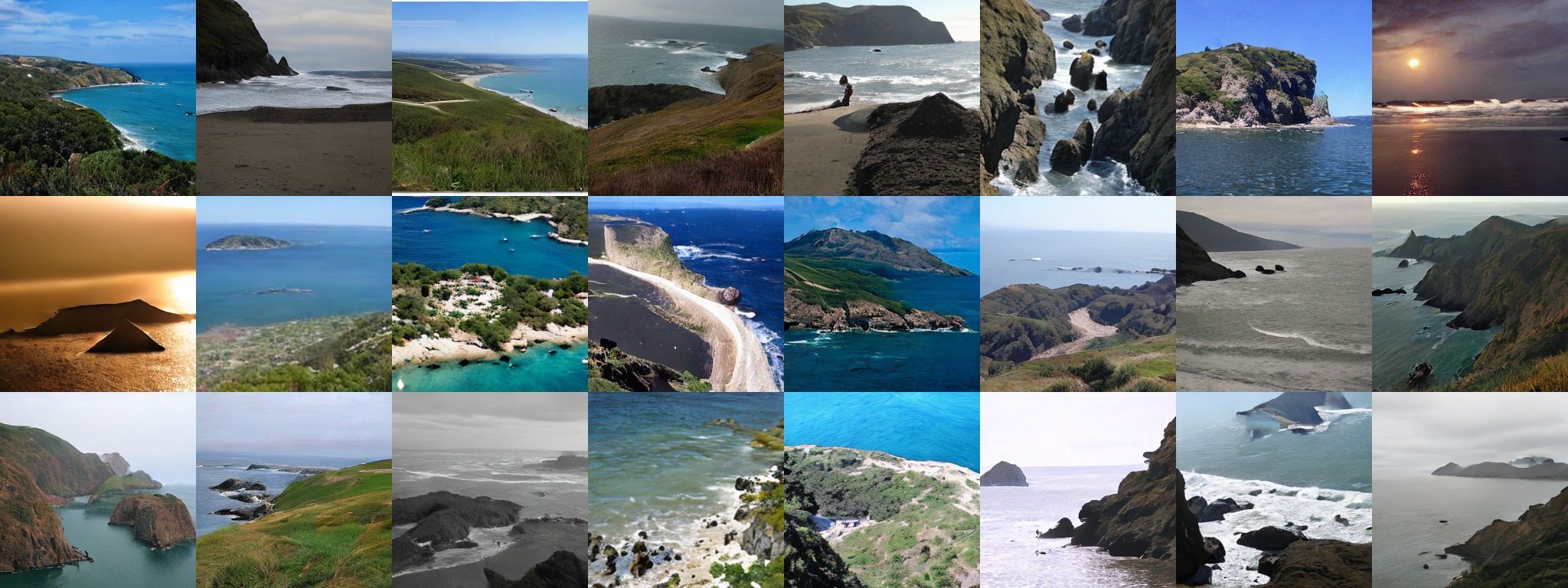}
        {\scriptsize class 976: promontory, headland, head, foreland}
        \vspace{.5em}
    \end{minipage}
    \hfill
    \begin{minipage}[t]{0.46\linewidth}
        \centering
        \includegraphics[width=\textwidth]{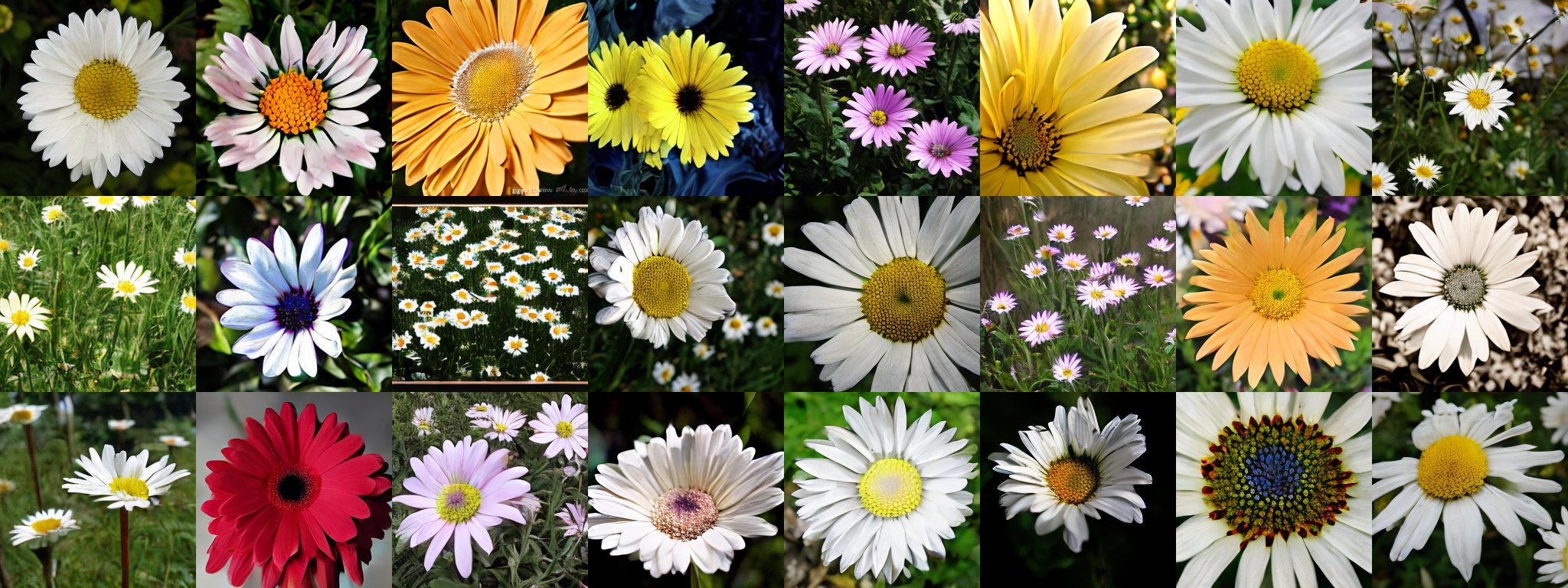}
        {\scriptsize class 985: daisy}
        \vspace{.5em}
    \end{minipage}
    \vspace{-1em}
    \caption{\textit{Uncurated} 1-NFE class-conditional generation samples of iMF-XL/2 on ImageNet 256$\times$256.}
    \vspace{-1em}
    \label{fig:appendix_gen2}
\end{figure*}